\documentclass{article}

\usepackage{microtype}
\usepackage{graphicx}
\usepackage{booktabs} 

\usepackage{hyperref}

\usepackage[accepted]{icml2024}

\usepackage{amsmath}
\usepackage{amssymb}
\usepackage{mathtools}
\usepackage{amsthm}

\usepackage{multicol}
\usepackage{wrapfig}

\usepackage[capitalize,noabbrev]{cleveref}

\theoremstyle{plain}

\theoremstyle{definition}

\theoremstyle{remark}

\usepackage[textsize=tiny]{todonotes}

\usepackage{multirow}
\usepackage{bm}
\usepackage{bbm}
\usepackage{wrapfig}
\usepackage{xcolor}
\usepackage{color, colortbl}
\usepackage{subcaption}

\newcommand*\rot{\rotatebox{90}}

\newcommand{\ie}{\textit{i.e. }}
\newcommand{\eg}{\textit{e.g. }}
\definecolor{gg}{gray}{0.70}

\definecolor{baselinecolor}{gray}{.9}
\newcommand{\cc}[1]{\cellcolor{baselinecolor}{#1}}

\icmltitlerunning{Cross-view Masked Diffusion Transformers for Person Image Synthesis}

\begin{document}

\twocolumn[
\icmltitle{Cross-view Masked Diffusion Transformers for Person Image Synthesis}



\icmlsetsymbol{equal}{*}

\begin{icmlauthorlist}
\icmlauthor{Trung X. Pham}{equal}
\icmlauthor{Zhang Kang}{equal}
\icmlauthor{Chang D. Yoo}{}\\
Korea Advanced Institute of Science and Technology (KAIST)
\end{icmlauthorlist}


\icmlcorrespondingauthor{Trung X. Pham}{trungpx@kaist.ac.kr}
\icmlcorrespondingauthor{Zhang Kang}{zhangkang@kaist.ac.kr}
\icmlcorrespondingauthor{Chang D. Yoo}{cd\_yoo@kaist.ac.kr}



\icmlkeywords{Machine Learning, ICML, XMDPT, mask diffusion transformers, human image synthesis}

\vskip 0.3in
]



\printAffiliationsAndNotice{\icmlEqualContribution} 

\begin{abstract}
We present X-MDPT (\underline{Cross}-view \underline{M}asked \underline{D}iffusion \underline{P}rediction \underline{T}ransformers), a novel diffusion model designed for pose-guided human image generation. X-MDPT distinguishes itself by employing masked diffusion transformers that operate on latent patches, a departure from the commonly-used Unet structures in existing works. The model comprises three key modules: 1) a denoising diffusion Transformer, 2) an aggregation network that consolidates conditions into a single vector for the diffusion process, and 3) a mask cross-prediction module that enhances representation learning with semantic information from the reference image. X-MDPT demonstrates scalability, improving FID, SSIM, and LPIPS with larger models. Despite its simple design, our model outperforms state-of-the-art approaches on the DeepFashion dataset while exhibiting efficiency in terms of training parameters, training time, and inference speed. Our compact 33MB model achieves an FID of 7.42, surpassing a prior Unet latent diffusion approach (FID 8.07) using only $11\times$ fewer parameters. Our best model surpasses the pixel-based diffusion with $\frac{2}{3}$ of the parameters and achieves $5.43 \times$ faster inference. The code is available at \url{https://github.com/trungpx/xmdpt}.
\end{abstract}

\begin{figure}
  \centering
  \includegraphics[width=1.0\linewidth]{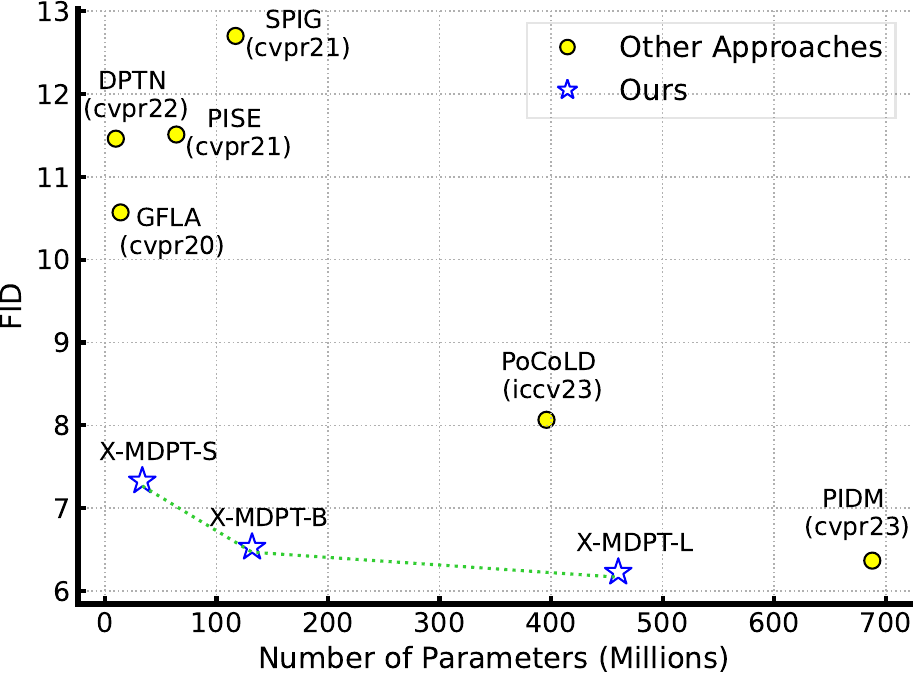}
  \vspace{-12pt}
  \caption{FID score of SOTAs approaches on the DeepFashion dataset. Our transformer-based models, X-MDPT (size of S, B, L) are marked in stars. X-MDPT-S surpasses the latent Unet-based PoCoLD with only $11\times$ fewer parameters.
  }
  \label{fig:fid_param}
  \vspace{-16pt}
\end{figure}

\vspace{-18pt}
\section{Introduction}
\label{intro_sec}
The task of Pose-guided Human Image Generation (PHIG) \cite{ma2017pose} has gained considerable attention with the advent of diffusion models recently. Initially, GAN-based approaches \cite{ma2017pose, men2020controllable, wu2023pose}, showed potential in PHIG but often struggled to generate target images accurately, resulting in high Frechet Inception Distance (FID) scores due to the presence of undesired artifacts. To address these challenges, \citet{bhunia2023person} introduced PIDM, a diffusion-based methodology employing iterative generation processes. While PIDM achieved state-of-the-art results in synthesizing high-quality images from target poses and source images, it suffered from slow inference speeds and high memory consumption owing to its pixel-based operation. In response to these efficiency concerns, \citet{han2023controllable} proposed PoCoLD, a latent diffusion framework operating on autoencoder latent outputs. However, while PoCoLD offered improvements in efficiency over PIDM, it fell short in terms of FID and Structural Similarity Index Metric (SSIM) metrics compared to PIDM. Notably, prevailing methodologies predominantly relied on Unet-based architectures with Convolutional Neural Networks (CNNs) for denoising diffusion processes. 

In contrast, our approach introduces a novel class of diffusion models based on transformers, aimed at addressing the challenges in PHIG more effectively. Diffusion models have demonstrated success in learning data distributions across various domains, particularly in generation tasks \cite{ho2020denoising, rombach2022high}. One significant advantage of diffusion models lies in their stable training process compared to Generative Adversarial Networks (GANs), which are prone to mode collapse \cite{isola2017image}. While Unet-based diffusion models, including Stable Diffusion \cite{rombach2022high} and related works \cite{zhang2023adding, zhao2023uni}, have achieved notable success in various applications, DiT \cite{peebles2023scalable} showcased the effectiveness of transformer-based designs in learning diffusion processes, effectively challenging traditional Unet-based networks in image generation. The emergence of masked diffusion transformers \cite{gao2023masked} as a state-of-the-art approach for class-conditional image generation on ImageNet further inspired our work. Motivated by these advancements, we propose a mask-based framework conditioned on both pose and style images to generate target images in the PHIG task \cite{ma2017pose, wu2023pose}. This novel approach aims to leverage semantic information from reference images, enhancing the model's ability to generate realistic and contextually coherent human images.

\begin{figure}[!htbp]
  \centering
  \includegraphics[width=1.0\linewidth]{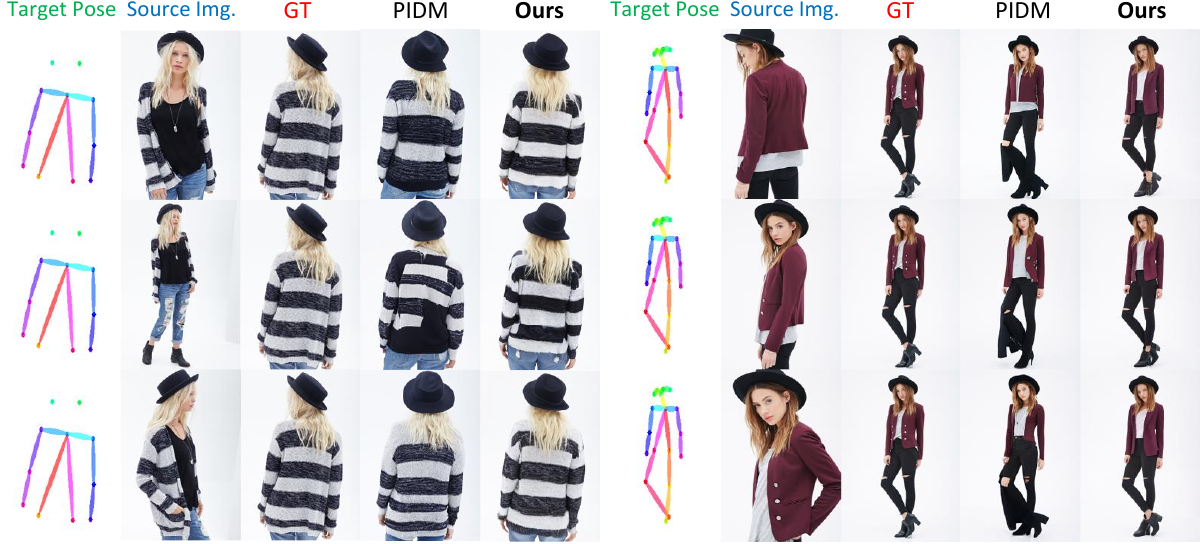}
  \vspace{-16pt}
  \caption{
  \textbf{Source-View Invariant}. The 2$^{\text{nd}}$ and 7$^{\text{th}}$ columns display different views of the same individuals from the DeepFashion. PIDM yields inconsistent outputs if varying source image views, whereas ours produces consistent ones closer to the ground truth. Best view at 200\% zoom.}
  \label{fig:source_invariant}
\vspace{-16pt}
\end{figure}

In this paper, we introduce X-MDPT, leveraging diffusion Transformer models for pose-guided human image generation. We design a specialized aggregation network to consolidate all conditions into a single vector to guide the diffusion Transformer via Adaptive Layernorm modulation. Additionally, we introduce a novel masking network to enhance the learning capabilities of transformers. By scaling up the number of layers or heads, we obtain configurations X-MDPT-S, X-MDPT-B, and X-MDPT-L, following standard transformer configurations \cite{dosovitskiy2010image, vaswani2017attention}. Despite its simple design, X-MDPT achieves state-of-the-art FID, SSIM, and LPIPS scores on the DeepFashion dataset while employing significantly fewer parameters compared to existing SOTAs (Fig. \ref{fig:fid_param}). X-MDPT combines simplicity and effectiveness, maintaining the scalability and flexibility of the Transformer architecture, demonstrating its potential for human image generation. Moreover, Fig. \ref{fig:source_invariant} demonstrates that our model produces more consistent and plausible outputs than the SOTA PIDM. Our contributions are as follows:

\begin{itemize}
    \vspace{-8pt}
    \item We propose X-MDPT, the first masked diffusion transformer framework for the PHIG task. Our model is scalable and efficient as it works on latent patches.
    \vspace{-6pt}
    \item We propose CANet to aggregate all conditions into a unified vector and we show that a single vector provides sufficient information to guide the diffusion process. This is conceptually simple as it neither requires any modification to the main diffusion framework nor extracts multiple-level features from different layers of conditions as did in existing approaches (CASD, PIDM, and PoCoLD).
    \vspace{-6pt}
    \item We propose a novel cross-view strategy to predict masked tokens across images, enhancing representation learning and improving generation quality.
    \vspace{-6pt}
    \item X-MDPT outperforms other state-of-the-art approaches on the DeepFashion dataset while being more efficient. For example, our compact models outperform the latent diffusion-based PoCoLD and pixel-based PIDM using only $11\times$ and $\frac{2}{3}\times$ fewer parameters, respectively. We qualitatively show that our model is robust to various difficult cases where the other approaches failed to generate the desired images.
\end{itemize}
\vspace{-8pt}

\section{Related Works}
\label{related_sec}
\textbf{Pose-guiged Person Image Synthesis.}
\citet{bhunia2023person} introduced PIDM as a solution to the PHIG problem using a diffusion model in pixel space, showing significant performance compared to traditional GAN-based methods. However, PIDM requires substantial computational resources and exhibits inefficiencies in both training and inference. \cite{han2023controllable} introduced PoCoLD, a latent diffusion model based on Unet design, to address these limitations. Concurrently, \citet{karras2023dreampose} presented DreamPose, applying the diffusion model to the fashion video domain by fine-tuning a text-to-image model, specifically Stable Diffusion \cite{rombach2022high}. While these approaches enhance the diffusion model's capabilities for PHIG with CNN-based denoisers, the potential of transformer-based methods remains untapped. To bridge this gap, we propose X-MDPT, leveraging pure transformer diffusion models to generate target person images.

\textbf{Diffusion Transformers.}
The CNN U-Net structure \cite{ronneberger2015u} initially served as the foundation for diffusion models and remains a standard choice across various diffusion-based generation tasks \cite{ho2020denoising, ho2022cascaded, song2019generative}. The introduction of DiT \cite{peebles2023scalable} marked a significant advancement, integrating the architecture of the pure transformer ViT \cite{dosovitskiy2021an} into latent diffusion models. DiT demonstrated exceptional scalability and outperformed Unet-like architectures. \citet{gao2023masked} further advanced the diffusion transformer model, achieving state-of-the-art class image generation on ImageNet by leveraging contextual representation learning. While they explored the potential of transformers in general generation tasks, our focus lies on the application of the diffusion transformer specifically to pose-guided human image generation (PHIG) within the Fashion domain for the first time. PHIG represents a pivotal and intricate generation task, requiring comprehensive information extraction from the source image including clothing, identity, background, and more, to faithfully generate the desired target pose \cite{ma2017pose, bhunia2023person}.

\textbf{Mask Prediction Modeling.}
Mask-based vision models, inspired by mask language models like BERT \cite{devlin2018bert}, have demonstrated remarkable scalability and performance. Notable examples include MAE \cite{he2022masked} in self-supervised learning (SSL), MaskGIT \cite{chang2022maskgit}, Muse \cite{chang2023muse}, and MAGVIT \cite{yu2023magvit} for learning discrete token distributions for image generation. In contrast, MDT \cite{gao2023masked} introduced an asymmetric masking schedule to enhance contextual representation learning in diffusion transformers. However, MDT struggles to establish correspondence between source and target images, limiting its representation learning capabilities. To address this limitation, we propose the Mask Inter-Prediction Network (MIPNet) inspired by prior works such as SiamMAE \cite{gupta2023siamese} and PatchMAE \cite{zhang2023patch} in SSL. MIPNet focuses on inter-semantic masking prediction within diffusion models for PHIG tasks. While it shares similarities with these MAEs in utilizing different views to predict masks, MIPNet differs in its lightweight, single self-cross-attention block and asymmetric masking, as opposed to these MAEs. Additionally, MIPNet is designed for generation tasks, while SimMAE/PatchMAE is for downstream recognition tasks.

\vspace{-6pt}
\section{Method}
\label{method_sec}
\begin{figure*}[!htbp]
  \centering
  \includegraphics[width=0.96\linewidth]{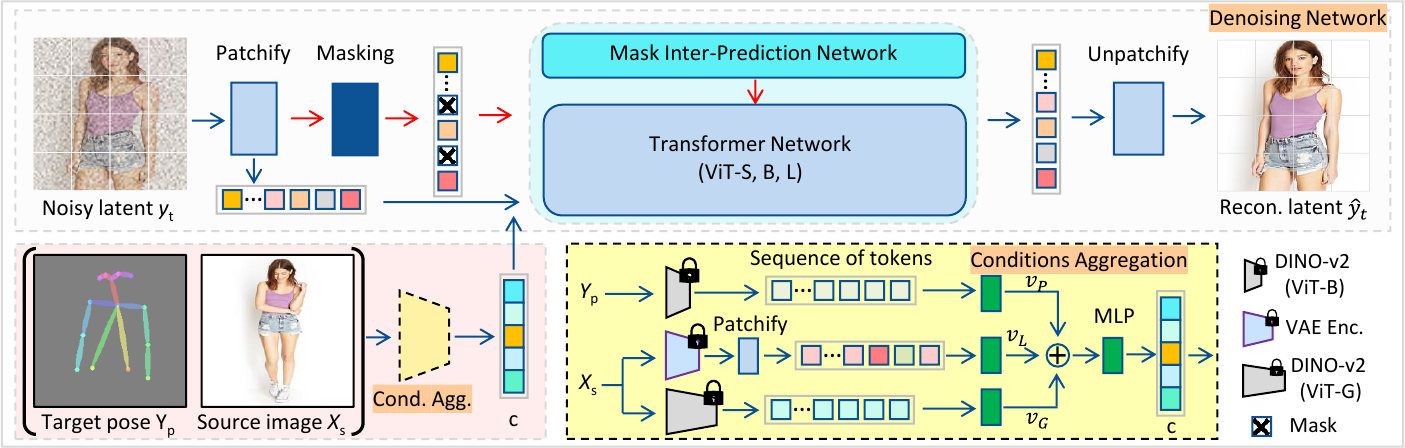}
  \vspace{-8pt}
  \caption{
  Overview of Our X-MDPT framework, built on transformers, facilitates pose-guided human image generation. During training, we randomly mask target image tokens at a 30\% ratio. The noisy target image is then processed through the Transformer Diffusion network, conditioned on the aggregated vector (with $D=768$ for our X-MDPT-B model) via AdaLN modulation \cite{peebles2023scalable}. Concurrently, we train a mask prediction objective alongside our novel mask inter-prediction network to capture semantics between source and target images when predicting mask tokens. 
  The \textcolor{red}{red arrow} $\color{red}\rightarrow$ signifies the training-only branch, discarded during inference, while the \textcolor{blue}{bue arrow}$\color{blue}\rightarrow$ serves both training and inference purposes. ``Cond. Agg.'' denotes ``Conditions Aggregation'' on the bottom right. VAE is omitted for simplicity.
  }
  \label{fig:arch}
  \vspace{-12pt}
\end{figure*}

We aim to design a simple yet scalable framework for addressing the PHIG task using Transformers. The overall architecture is depicted in Fig. \ref{fig:arch}. Our X-MDPT comprises three core modules: 1) Transformer-based Denoising Diffusion Network (TDNet), 2) Conditional Aggregation Network (CANet), and 3) Mask Inter-Prediction Network (MIPNet). Here, TDNet performs the denoising diffusion, while CANet consolidates all necessary condition inputs into a single vector for TDNet's input. Additionally, MIPNet enhances the diffusion learning process by predicting masked tokens using a novel reference-based predictor. We provide detailed insights into each component next.

\vspace{-6pt}
\subsection{Transformer-based Denoising Diffusion Network}
In our X-MDPT framework, we denote this network component as TDNet for brevity. TDNet is built on top of DiT \cite{peebles2023scalable} to establish the diffusion process using Transformer architecture. Here's an overview of the framework for a $256\times 256$ resolution case: Given the source image $X_s\in\mathbb{R}^{256\times 256\times 3}$ and the target pose $y_p$, the objective is to learn the model parameterized by $\theta$ to capture the target pose and the style of the source image to generate the final target image $Y\in\mathbb{R}^{256\times 256\times 3}$. Initially, we employ a pre-trained VAE \cite{rombach2022high} to map the pixel images to latent representations $x_s\in\mathbb{R}^{32\times 32\times 4}$ and $y\in\mathbb{R}^{32\times 32\times 4}$ for denoising. The denoising network $\epsilon_{\theta}$ is a transformer-based diffusion model that learns the condition distribution $p_\theta(y|x_s, y_p)$. The denoising process progressively adds Gaussian noise $\epsilon\sim \mathcal{N}(0,\mathbf{I})$ to image $y$ to obtain $y_t$ at timestep $t\in [1, T]$. The conditions $x_s$ and pose $y_p$ are represented by $c$. The training objective is to predict the added noise using mean squared error:
\begin{equation}
    \mathcal{L_\text{denoise}} = \mathbb{E}_{y,c,\epsilon\sim\mathcal{N}(0,\mathbf{I}),t}\Vert \epsilon - \epsilon_\theta(y_t,c,t) \Vert^2.
\end{equation}
Once $p_\theta$ is trained, inference proceeds by initiating with a random noise image $y_{T}\sim \mathcal{N}(0,\mathbf{I})$ and iteratively sampling $y_{t-1}\sim p_\theta(y_{t-1}|y_t)$ to obtain the final target image $y_0$. Our diffusion transformer network adheres to the structure outlined in DiT \cite{peebles2023scalable}. We transform the noisy latent $y_t\in \mathbb{R}^{32\times 32\times 4}$ into patches with a patch size of $p=2$, forming the sequence $z_{y_t}=[z_y^{(1)}, z_y^{(2)}, ..., z_y^{(L_y)}]\in \mathbb{R}^{L_y\times D}$, where $L_y$ and $D$ denote the sequence length and dimension, respectively. The condition $c$ is integrated into the TDNet through adaptive layer normalization (AdaLN-Zero), following the default setup of DiT.

\vspace{-6pt}
\subsection{Conditional Aggregation Network}
The CANet network integrates three inputs: 1) the target pose condition feature (TPF), 2) the local source image feature (LSIF) obtained from the output of the VAE, and 3) the global source image feature (GSIF) derived from the pre-trained feature of DINOv2. CANet processes these inputs to produce a unified vector $c$ with dimensions matching the width of the Transformer.

\textbf{Local Source Image Feature.}
This feature ensures alignment with the noisy target image within the TDNet, enabling the transfer of information from the source image (including clothing, person, and background) to generate the target image. Specifically, the latent image $32\times 32\times 4$ is converted into a sequence $z_{x_s}=[z_x^{(1)}, z_x^{(2)}, ..., z_x^{(L_x)}]\in \mathbb{R}^{L_x\times D}$, where $L_x = 256$ tokens ($\text{LSIF} = z_{x_s} \in \mathbb{R}^{256\times D}$). An $1\times 1$ conv. layer is then applied to linearly map 256 channels to a single channel, yielding the local vector $v_L \in \mathbb{R}^{D}$.

\textbf{Pose Representation.}
We utilize a 3-channel RGB visualization image of the pose ($256\times 256\times 3$), resized to $224\times 224\times 3$, as the input to CANet. In contrast, PIDM \cite{bhunia2023person} employs a more complex pose representation with $256\times 256\times 20$, where the 20 channels are 3 RGB channels and 17 Gaussian heatmaps. Our experiments demonstrate that the use of a simple RGB pose representation is adequate for generating satisfactory person images. We employ pre-trained DINOv2-B to extract RGB pose features, resulting in a CLS and 256 patch tokens, concatenated into a sequence of 257 ($\text{TPF} \in \mathbb{R}^{257\times D}$), which is then processed by one $1\times 1$ convolutional layer to map 257 channels to 1 channel, yielding vector $v_P \in \mathbb{R}^{D}$.

\textbf{Global Source Image Feature.}
We observed that relying solely on local features from the source image is insufficient for the Transformer to capture both the details and the identity of the person. As demonstrated in AnyDoor \cite{chen2023anydoor}, self-supervised models like DINO features can effectively capture identity and details. Hence, we utilize DINOv2-G to extract the CLS token and concatenate all other tokens to form the global feature ($\text{GSIF} \in \mathbb{R}^{257\times D}$). For a resolution of $256\times 256$, we resize it to $224\times 224$ as required by DINO (where both width and height are divisible by 14) to obtain a CLS token and 256 patch tokens. Interestingly, for a resolution of $512\times 512$, we find that using the same resolution of $224\times 224$ for both pose and global features when extracting DINO tokens is effective, eliminating the need for extracting pose and global features of size $448\times 448$ (close to $512\times 512$). This helps save memory, as the output feature sequence of the DINO transformer is larger in this case ($32\times 32=1024$) compared to $16\times 16=256$ tokens (DINO output for $224\times 224$ images). This approach differs from Unet-based frameworks like PIDM and PoCoLD, which necessitate exact $512\times 512$ size for the pose condition. Finally, we pass the resulting sequence $\text{GSIF} \in \mathbb{R}^{257\times D}$ through one $1\times 1$ conv. layer to obtain the global vector $v_G \in \mathbb{R}^{D}$.

\textbf{Aggregation.} We obtain the final vector by either simply using an addition operation or concatenating them to have three channels and use a $1\times 1$ convolution operation $\mathcal{H}$ to get a unified conditional vector c, \ie $c = \mathcal{H}(v_L, v_P, v_G) \in \mathbb{R}^D$. We find that an MLP gives a slightly better FID as this MLP can automatically put weights on each condition and is learned during backpropagation. 
We show that having both local and global vectors, \ie $v_L + v_G$ of the source image together with $v_P$ is crucial to generating the high-quality target image in the ablation section.

\subsection{Mask Inter-Prediction Network}
\begin{figure}
  \centering
  \includegraphics[width=1.0\linewidth]{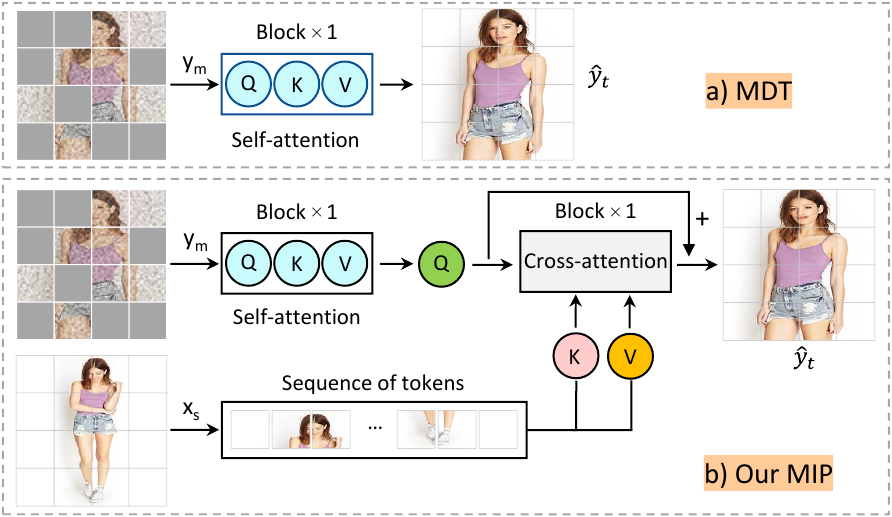}
  \vspace{-16pt}
  \caption{
  \textbf{MIPNet vs. MDT}. Ours MIPNet predicts masked tokens by using all tokens from the reference image $x_s$. 
  }
  \label{fig:MIPNet}
\vspace{-16pt}
\end{figure}

\citet{gao2023masked} demonstrated improvements in transformer-based diffusion models by introducing a lightweight predictor to fill masked regions within images, resulting in enhanced FID scores. While this masking strategy proved effective in general image-generation tasks like ImageNet, its application to the DeepFashion dataset remained unexplored. In the PHIG task, merely predicting masks using unmasked tokens within an image falls short of capturing the necessary correspondence between source and target images, critical for conditional generation. As a result, this approach yields suboptimal performance in person image generation. To address this limitation, we propose a novel prediction module that integrates information from the source image (cross-view) to guide the diffusion model in predicting masked tokens within the target image. This is in contrast to MDT's reliance solely on the target image. By incorporating information from the reference image, our approach, MIPNet, gains richer contextual cues for completing mask patches in the target image and learning meaningful semantic correspondence. The primary distinction between MDT and MIPNet is illustrated in Fig. \ref{fig:MIPNet}. In the ablation section, we validate the effectiveness of our method compared to MDT's mask prediction. The objective loss for training with masked tokens remains consistent with the standard loss $\mathcal{L}_{\text{denoise}}$.
\begin{equation}
    \mathcal{L}_{\text{mask}} = \mathbb{E}_{y,c,\epsilon\sim\mathcal{N}(0,\mathbf{I}),t}\Vert \epsilon - \epsilon_\theta(f_\theta(x_s, y_m),c,t) \Vert^2,
\end{equation}
where $f_\theta$ is the network that contains MIPNet $g_\theta$, the $N_1$ encoder layers and $N_2$ decoder layers of the TDNet. Here, $N_1$ and $N_2$, along with other settings, remain consistent with those defined in MDT \cite{gao2023masked}. The output of MIPNet is computed by the following equation:
\begin{align}
\begin{split}
    \label{mipnet}
     g_\theta(z_{x_s}, z_{y_m}) &= \phi_{\text{attn}}(z_{y_m}, z_{y_m}, z_{y_m}) \\
    & + \phi_{\text{attn}} \left( \phi_{\text{attn}}(z_{y_m}, z_{y_m}, z_{y_m}), z_{x_s}, z_{x_s} \right),
    \end{split}
\end{align}
where $\phi_\text{attn}$ denotes the attention mechanism proposed by \cite{vaswani2017attention} to learn cross-view mask prediction:
\begin{equation}
    \label{attention}
    \phi_\text{attn}(\mathbf{Q}, \mathbf{K}, \mathbf{V}) = \text{Softmax} \left( \frac{\mathbf{QK}^\top}{\sqrt{d_k}} \right) \mathbf{V}
\end{equation}

\textbf{Final objective function.} We jointly optimize two objective functions in parallel through the following equation:
\begin{equation}
\label{eq:final}
    \mathcal{L_\text{total}} = \mathcal{L}_{\text{denoise}} + \mathcal{L}_{\text{mask}}.
\end{equation}
In Eq. \ref{eq:final}, if $\mathcal{L}_{\text{mask}}$ is removed, the framework resembles the style of DiT. Alternatively, replacing $g_\theta$ with only the self-attention transforms the model into the MDT style. During inference, MIPNet is omitted, retaining only positional embeddings, as implemented in MDT \cite{gao2023masked}.

\subsection{Classifier-Free Guidance}
We use the common technique of classifier-free guidance \cite{ho2022classifier} to predict noise via the linear combination of the unconditional model $\epsilon_\theta(y_t,t)$ and conditional model $\epsilon_\theta(y_t,c,t)$ as follows:
\begin{equation}
    \hat{\epsilon}_\theta(y_t,x,t) = \gamma_t \epsilon_\theta(y_t,c,t) + (1-\gamma_t)\epsilon_\theta(y_t,t).
\end{equation}
The guidance scale $\gamma_t$ is determined at timestep $t$. During training, we randomly assign the unified conditional vector $c\in\mathbb{R}^D$, obtained by CANet, to the zero vector $\emptyset\in\mathbb{R}^D$ with a probability of $\eta=10\%$. Another parameter that facilitates dynamic scale guidance is $\gamma_t=\frac{1-\cos \pi(\frac{t}{T})^\alpha}{2}\times \gamma$. This sets the power-cosine schedule ($\alpha$) for the guidance scale during the sampling procedure, as used in MDT \cite{gao2023masked}. By default, we set $\gamma=2.0$. Our experiments indicate that using $\alpha=1.0$ yields the best FID, albeit slightly lower values for other metrics such as SSIM and LPIPS. Conversely, $\alpha=0.01$ results in slightly higher FID but improved SSIM and LPIPS.

\vspace{-8pt}
\section{Experiments}
\label{experiments_sec}
\subsection{Implementation Details}
\textbf{Dataset.}
We evaluate our method against the state-of-the-art (SOTA) using high-resolution images from the DeepFashion In-shop Clothes Retrieval Benchmark dataset \cite{liu2016deepfashion} at resolutions of $256\times 256$ and $512\times 512$. The dataset comprises non-overlapping train and test subsets, containing 101,966 and 8,570 pairs, respectively. We adopt preprocessing steps consistent with prior works \cite{bhunia2023person, han2023controllable}. We employ OpenPose \cite{cao2017realtime} to extract 18 key points from each person's image and then utilize OpenCV to generate RGB visualizations, which are resized to $224\times 224$. This fixed size is applied for pose images in both the $256\times 256$ and $512\times 512$ resolution cases.
\textbf{Metrics.}
We use the common measurements as utilized in prior studies \cite{bhunia2023person, han2023controllable} including FID, SSIM, LPIPS, and optionally PSNR for ablations.

\textbf{Training.}
We finetune the pre-trained VAE ft-MSE of Stable Diffusion on the DeepFashion training set. For $256\times 256$ images, training was conducted on a single A100 GPU (80GB RAM) with a batch size of 32, spanning 800k steps. Meanwhile, for $512\times 512$ images, we employed two A100 GPUs with a batch size of 10 (5 images per GPU), trained for 1M steps. For ablations, we trained X-MDPT-B with 300k steps on a $256\times 256$ resolution. The learning rate was set to 1e-4, the model's EMA rate to 0.9999, and other settings aligned with DiT \cite{peebles2023scalable}. Note that, the original images in the DeepFashion dataset have resolutions of $256\times 176$ and $512\times 352$, which we resized to $256\times 256$ and $512\times 512$, respectively, using bicubic interpolation before inputting them into the models.

\subsection{Main Results}
\label{deepfashiion_result_sec}
Results are reported in Tab. \ref{tab:compare_deepfashion}, with the comparative performance of different approaches. Our X-MDPT demonstrates consistent superiority across FID, SSIM, and LPIPS metrics at a resolution of $256\times 256$. We find that our model achieves its best FID score, around 6.25, during the mid-training phase (350-400k steps). Subsequently, with extended training (800k-1M steps), the FID stabilizes at around 7.28, while SSIM and LPIPS metrics exhibit continual enhancement.
Here, the evaluation process performs the comparison between the synthetic test set and the real training data. Ground truth images yield an FID of 7.86, indicating a substantial distribution gap in DeepFashion's test and training sets, as confirmed by PoCoLD \cite{han2023controllable}. As the model fully converges, we observe a narrowing of the FID, LPIPS, and SSIM as the model learned generates images closer to the ground truth. At a higher resolution of $512 \times 512$, X-MDPT exhibits a slight lag behind PIDM in FID, but better in other key metrics. Its resource efficiency-batch sizes of 32 and 10 with one and two GPUs for 256 and 512 resolutions, suggests room for improvement compared to PIDM's 8 GPUs with a batch size of 128.

\vspace{-6pt}
\begin{table}[!htbp]
    \caption{Comparison of X-MDPT with SOTA approaches. $\dag$ we reproduced with the public checkpoint. \textbf{Bold} and \underline{undeline} denotes the best and second-best, respectively.}
    \vspace{-10pt}
    \label{tab:compare_deepfashion}
    \begin{center}
    \resizebox{1.0\hsize}{!}{
    \begin{tabular}{cccccc}
    \toprule
    \bf Dataset &\bf  Method &\bf  FID $\downarrow$ &\bf  SSIM $\uparrow$ &\bf  LPIPS $\downarrow$ &\bf  Type \\
    \midrule
    \multirow{19}{*}{DeepFashion} & Def-GAN \cite{siarohin2018deformable} & 18.457 & 0.6786 & 0.2330 & \multirow{8}{*}{\rot{{\large Non-Diffusion}}} \\
    \multirow{19}{*}{($256 \times 176$)} & PATN \cite{zhu2019progressive} & 20.751 & 0.6709 & 0.2562 \\
    & ADGAN \cite{men2020controllable} & 14.458 & 0.6721 & 0.2283 \\
    & PISE \cite{zhang2021pise} & 13.610 & 0.6629 & 0.2059 \\
    & GFLA \cite{ren2020deep} & 10.573 & 0.7074 & 0.22341 \\
    & DPTN \cite{zhang2022exploring} & 11.387 & 0.7112 & 0.1931 \\
    & CASD \cite{zhou2022cross} & 11.373 & 0.7248 & 0.1936 \\
    & NTED \cite{ren2022neural} & 8.6838 & 0.7182 & 0.1752 \\
    \cmidrule{2-6}
    & PoCoLD \cite{han2023controllable} & 8.0667 & 0.7310 & \underline{0.1642} & \multirow{3}{*}{\rot{{\large Unet}}} \\
    & PIDM \cite{bhunia2023person} & \underline{6.3671} & \underline{0.7312} & 0.1678 &  \\
    & PIDM \cite{bhunia2023person}$\dag$ & 6.6182 & 0.7294 & 0.1715 &  \\
    \cmidrule{2-6}
    & \bf X-MDPT-S (300k), $\alpha=0.01$ & 7.4282 & 0.7128 & 0.1961 & \multirow{6}{*}{\rot{{\large Transformer}}} \\
    & \bf X-MDPT-S (800k), $\alpha=0.01$ & 7.6724 & 0.7194 & 0.1875 &  \\
    & \bf X-MDPT-B (300k), $\alpha=0.01$ & 6.7288 & 0.7215 & 0.1814 &  \\
    & \bf X-MDPT-B (800k), $\alpha=0.01$ & 7.3293 & 0.7284 & 0.1734 &  \\
    & \bf X-MDPT-L (350k), $\alpha=1.00$ & \bf 6.2512 & 0.7298 & 0.1671 & \\
    & \bf X-MDPT-L (800k), $\alpha=0.01$ & 7.2865 & \bf 0.7405 & \bf 0.1589 & \\
    \cmidrule{2-6}
    & \textcolor{gg}{VAE reconstruction} & \textcolor{gg}{8.0126} & \textcolor{gg}{0.9168} & \textcolor{gg}{0.0142} & \multirow{2}{*}{{\large - }} \\
    & \textcolor{gg}{Ground Truth} & \textcolor{gg}{7.8610} & \textcolor{gg}{1.0000} & \textcolor{gg}{0.0000} &  \\
    \midrule
    \multirow{7}{*}{DeepFashion} & CocosNet-v2 \cite{zhou2021cocosnet} & 13.325 & 0.7236 & 0.2265 & \multirow{1}{*}{\rot{ GAN }} \\
    \multirow{7}{*}{($512 \times 352$)} & NTED \cite{ren2022neural} & 7.7821 & 0.7376 & 0.1980 & \\
    & PoCoLD \cite{han2023controllable} & 8.4163 & \underline{0.7430} & 0.1920 & \multirow{1}{*}{\rot{ Unet }} \\
    & PIDM \cite{bhunia2023person} & \bf 5.8365 & 0.7419 & \underline{0.1768} & \\
    \cmidrule{2-6}
    & \bf X-MDPT-L (500k), $\alpha=1.0$ & \underline{5.9264} & 0.7416 & 0.1788 & \multirow{2}{*}{\rot{Trans}} \\
    & \bf X-MDPT-L (1M), $\alpha=0.01$ & 7.1615 & \bf 0.7522 & \bf 0.1645 & \\
    \cmidrule{2-6}
    & \textcolor{gg}{VAE reconstruction} & \textcolor{gg}{8.1815} & \textcolor{gg}{0.9122} & \textcolor{gg}{0.0266}  & \multirow{2}{*}{{\large - }}\\
    & \textcolor{gg}{Ground Truth} & \textcolor{gg}{7.9150} & \textcolor{gg}{1.0000} & \textcolor{gg}{0.0000}  & \\
    \bottomrule
    \end{tabular}}
    \end{center}
\end{table}
\vspace{-16pt}

\begin{figure*}[!ht] 
  \centering
  \includegraphics[width=1.0\linewidth]{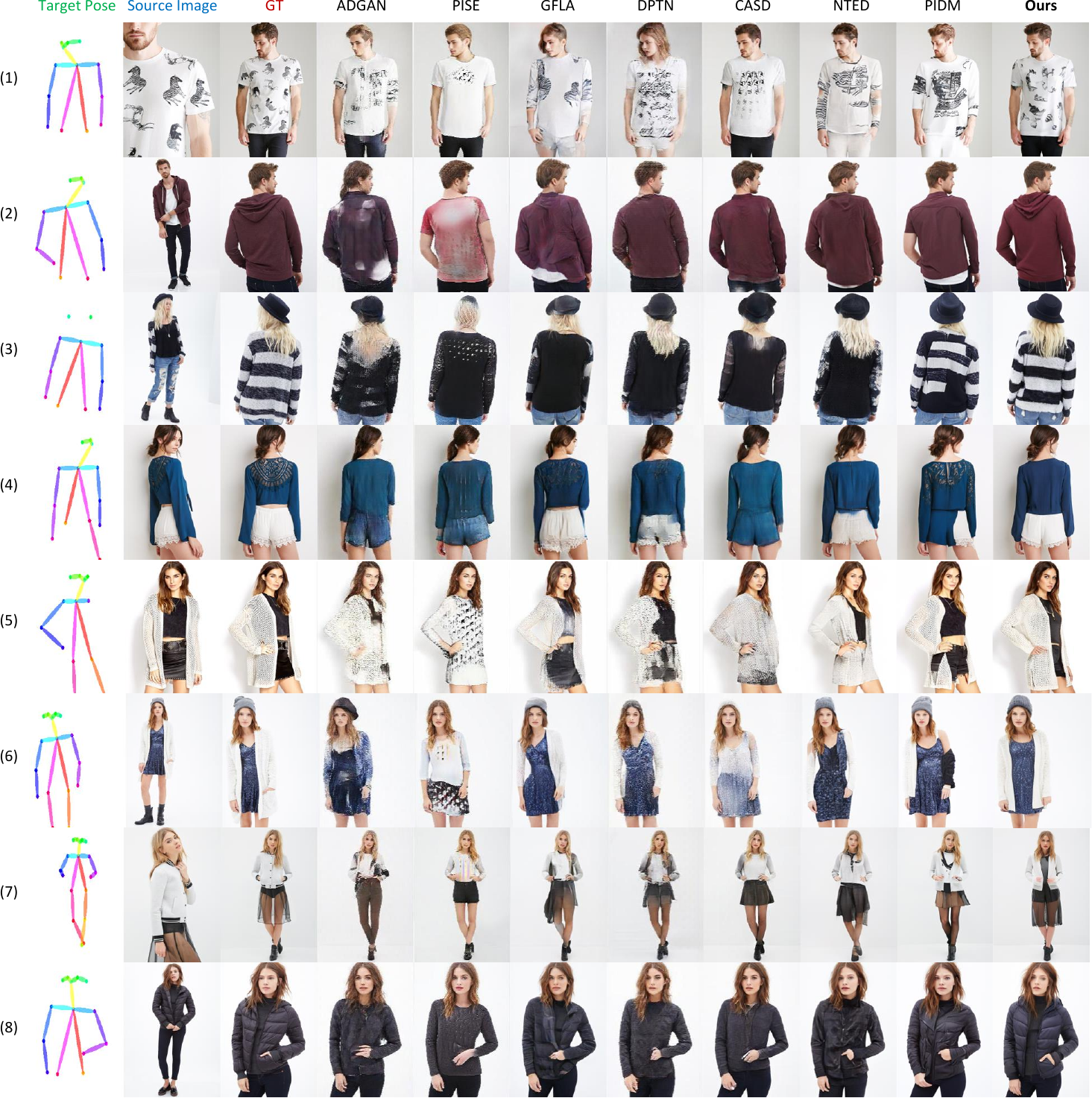}
  \vspace{-10pt}
  \caption{ \textbf{Qualitative comparison.} Person images are generated by state-of-the-art approaches on the DeepFashion dataset. Given two inputs: the \textcolor{green}{target pose} and \textcolor{blue}{source image}, our transformer-based X-MDPT-L (460M) handles various difficult cases and creates high-quality, more realistic, and closer to the ground truth (\textcolor{red}{GT}) compared to other CNN-based methods. }
  \label{fig:main_figure4}
  \vspace{-10pt}
\end{figure*}

\textbf{Qualitative Results.} We compared the output generated by X-MDPT and other methods in Fig. \ref{fig:main_figure4}. The generated person images exhibit high quality across various scenarios. Notably, pixel-based diffusion models like PIDM and other CNN-based methods often struggle to faithfully capture intricate style details, resulting in noticeable artifacts, as observed in specific cases. In contrast, X-MDPT operates on latent space with a transformer equipped with a semantic understanding scheme, enabling a more accurate depiction of clothing elements such as shirts and trousers. This results in more complete and satisfactory images that align well with the intended pose and source image.

Visual comparisons between ours and existing approaches in Fig. \ref{fig:main_figure4} highlight X-MDPT's consistent generation of plausible and complete target images that closely resemble ground truth. Furthermore, compared to the previous best-performing model, PIDM, our model demonstrates superior alignment when the image viewpoint is changed, as shown in Fig. \ref{fig:source_invariant}. We analyze this property in Fig. \ref{fig:cosine}. More qualitative results can be found in the \textbf{Appendix}.

\subsection{Ablation Studies}
\label{alation_sec}

\textbf{Learning Invariant Views of Source Images.} For the task of PHIG, with different views of a person's image and the same target pose, we expect to generate the same target image. However, as shown in Fig. \ref{fig:source_invariant}, PIDM does not capture the specific structure of the clothes and produces inconsistent target images when the person's view is changed. By contrast, X-MDPT gives consistent target images and it is closer to the ground truth. 
\begin{wrapfigure}{r}{0.6\linewidth}
  \vspace{-6pt}
  \centering
  \includegraphics[width=1.0\linewidth]{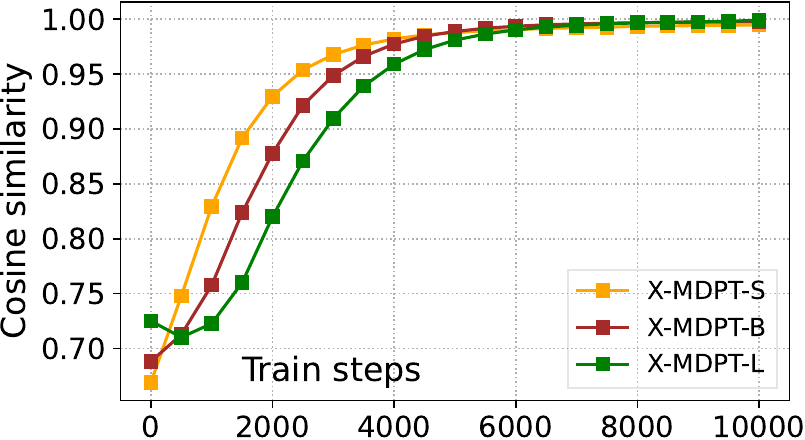}
  \vspace{-14pt}
  \caption{
  \textbf{CANet.} Views of same person have 99.99\% similarity.}
  \label{fig:cosine}
  \vspace{-6pt}
\end{wrapfigure}
This consistent generation can be elaborated by measuring the cosine similarity of unified conditional vectors when varying views of the source image. We find that existing works such as PoCoLD and PIDM utilize conditions in different places and multiple levels in their networks, which makes it challenging to produce the unified vectors of all conditions (\ie pose and source image). 

By contrast, our CANet module supports generating such a unified conditional vector. These vectors can easily support monitoring the cosine similarity score of different views of the same person. We conducted the whole test set of DeepFashion and took the average. Fig. \ref{fig:cosine} shows that the similarity reached above 99.99\% after 10k training steps, indicating that CANet learns to capture the invariant features of the same person. This explains why X-MDPT can generate a consistent target given the same pose with different views of a person, which is helpful for the task.

\begin{figure}[!htbp]
  \centering
  \includegraphics[width=1.0\linewidth]{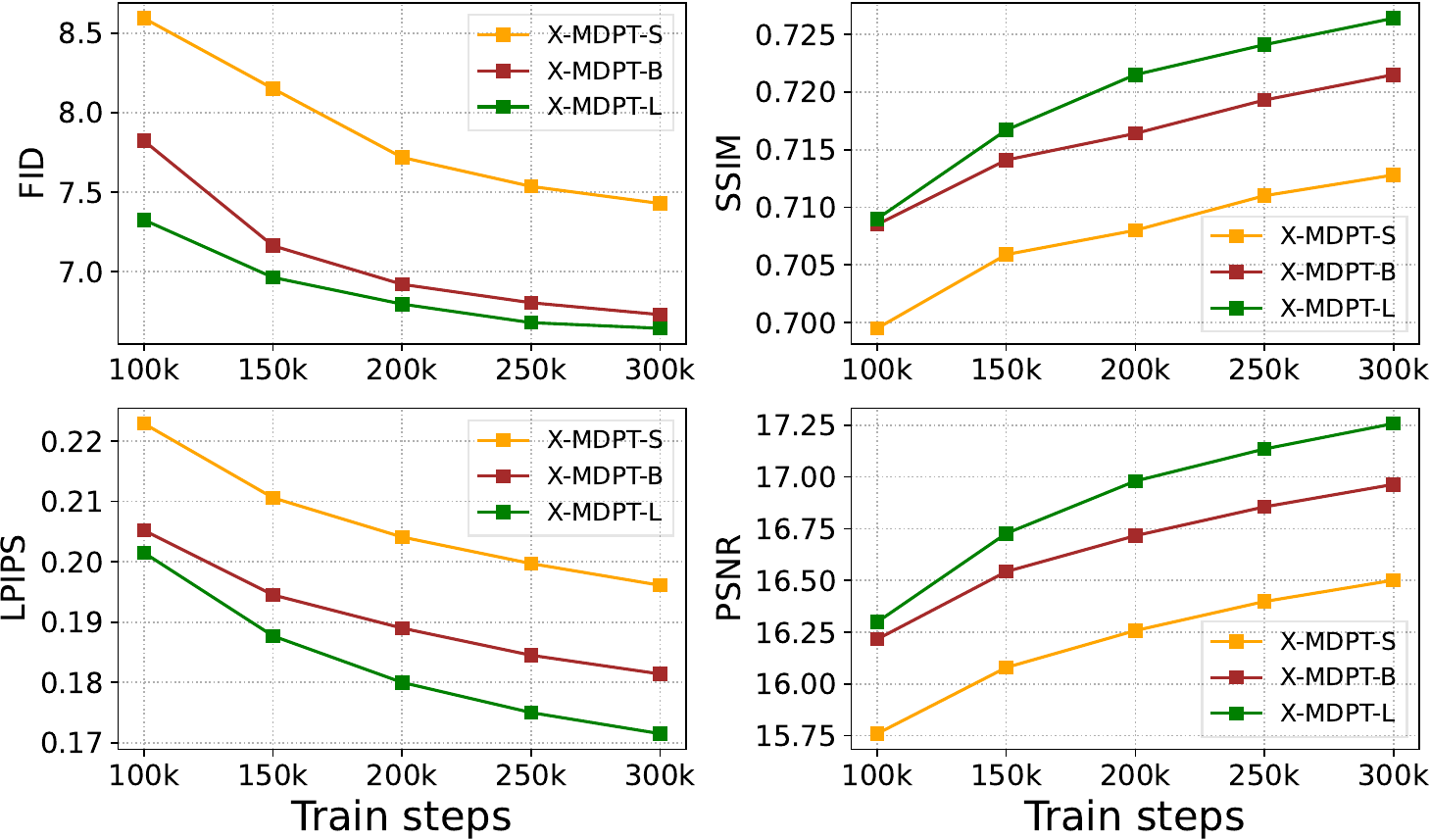}
  \vspace{-16pt}
  \caption{\textbf{Scalability.} X-MDPT (model sizes S, B, and L) is scalable for four metrics FID, SSIM, LPIPS, and PSNR.}
  \label{fig:ablation_scalability}
  \vspace{-10pt}
\end{figure}

\textbf{Scalability.} We discover the scalability of X-MDPT with sizes S, B, and L in Fig. \ref{fig:ablation_scalability}. We observe FID, SSIM, LPIPS, and PSNR improved as scaling up model size, demonstrating the potential of transformers for the PHIG problem.

  \vspace{-8pt}

\textbf{Impact of Mask Prediction.} Tab. \ref{tab:transformer_baselines} shows that naive applying transformer (DiT) \cite{peebles2023scalable} to PHIG task does not give a satisfactory FID. MDT \cite{gao2023masked} can help improve FID but is not beneficial for SSIM and LPIPS. By contrast, our method improves all metrics. 

\begin{figure}[!htbp]
  \centering
  \includegraphics[width=1.0\linewidth]{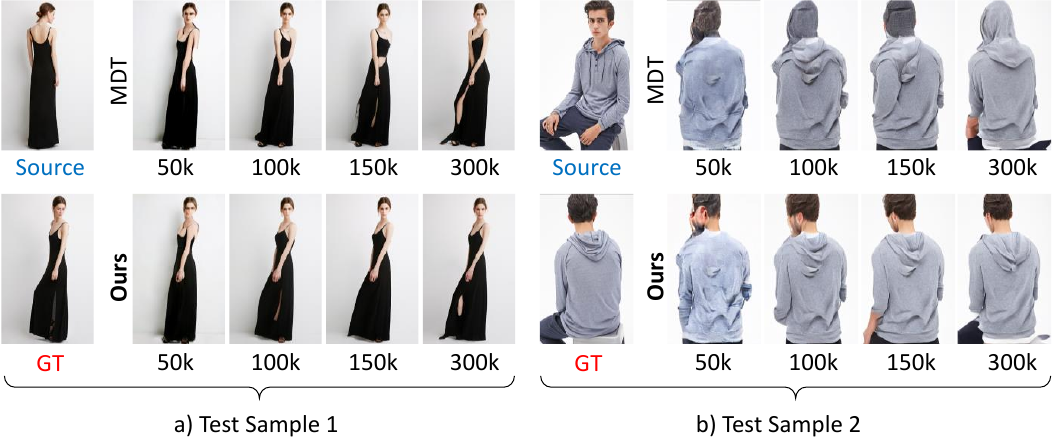}
  \vspace{-10pt}
  \caption{\textbf{Generated images with different training steps}. Our X-MDPT performs much better than MDT.
  It is best viewed with zoom in 200\%.}
  \label{fig:image_steps7}
  \vspace{-8pt}
\end{figure}

Qualitatively, images generated by our method are more realistic and closer to ground truth images as shown in Fig. \ref{fig:image_steps7}. This can be attributed to the fact that correspondence is the key to achieving the best performance in the posed-guide person image generation task, for which our X-MDPT equipped with the proposed Inter-Semantic module (MIPNet) can capture strong clues between the source and target images.

\begin{table*}[!htbp]
    \caption{ \textbf{Ablation experiments} on the DeepFashion dataset at $256\times 176$ resolution. Default settings are marked in \colorbox{baselinecolor}{gray}. }
    \vspace{-8pt}
    \label{tab:all_ablations}
    \centering
    \subfloat[\textbf{Transformer baselines}. Compared different diffusion transformers. MIPNet improves all metrics.
    \label{tab:transformer_baselines}]{
    \centering
    \begin{minipage}{0.29\linewidth}{
    \begin{center}
    \resizebox{1.0\hsize}{!}{
    \begin{tabular}{ccccc}
    Method & FID $\downarrow$ & SSIM $\uparrow$ & LPIPS $\downarrow$ & PSNR $\uparrow$  \\
    \midrule
    Transformer & 7.1150 & \underline{0.7199} & \underline{0.1841} & \underline{16.8882} \\
    + MDT & \underline{6.8616} & 0.7182 & 0.1851 & 16.8227 \\
    \bf + MIPNet & \bf \cc{6.7288} & \bf \cc{0.7215} & \bf \cc{0.1814} & \bf \cc{16.9642} \\
    \end{tabular}}
    \end{center}}
    \end{minipage}}
    \hfill
    \subfloat[\textbf{Masking ratio effect}. Model X-MDPT-B trained on the DeepFashion dataset for 300k iterations.
    \label{tab:masking}]{
    \centering
    \begin{minipage}{0.288\linewidth}{
    \begin{center}
    \resizebox{1.0\hsize}{!}{
    \begin{tabular}{ccccc}
    Mask Ratio & FID $\downarrow$ & SSIM $\uparrow$ & LPIPS $\downarrow$ & PSNR $\uparrow$ \\
    \midrule
    30\% & \bf \cc{6.7288} & \cc{0.7215} & \cc{0.1814} & \cc{16.9642} \\
    50\% & \underline{6.8081} & \underline{0.7231} & \underline{0.1792} & \underline{17.0609} \\
    70\% & 6.9533 & \bf 0.7246 & \bf 0.1759 & \bf 17.1442 \\
    \end{tabular}}
    \end{center} }
    \end{minipage} }
    \hfill
    \subfloat[\textbf{Global \& pose feature}. (G), (B), and (S) mean DINOv2-G, DINOv2-B, and DINOv2-S, respectively. 
    \label{tab:global}]{
    \centering
    \begin{minipage}{0.34\linewidth}{
    \begin{center}
    \resizebox{1.0\hsize}{!}{
    \begin{tabular}{ccccc}
    Features & FID $\downarrow$ & SSIM $\uparrow$ & LPIPS $\downarrow$ & PSNR $\uparrow$  \\
    \midrule
    Global (S) + Pose (S) & 6.9239 & 0.7187 & 0.1894 & 16.7466 \\
    Global (G) + Pose (S) & 6.9237 & \bf 0.7228 & \bf 0.1798 & \bf 17.011 \\
    Global (G) + Pose (B) & \bf \cc{6.7288} & \underline{\cc{0.7215}} & \underline{\cc{0.1814}} & \underline{\cc{16.9642}} \\
    \end{tabular}}
    \end{center} }
    \end{minipage} }
    \vspace{8pt}
    \subfloat[\textbf{Attention for MIPNet}. Compared performance of different designs, \textbf{self-cross} attention gives the best FID.
    \label{tab:attention_ablation}]{
    \centering
    \begin{minipage}{0.29\linewidth}{
    \begin{center}
    \resizebox{1.0\hsize}{!}{
    \begin{tabular}{ccccc}
    Method & FID $\downarrow$ & SSIM $\uparrow$ & LPIPS $\downarrow$ & PSNR $\uparrow$  \\
    \midrule
    self-att & 6.8616 & 0.7182 & 0.1851 & 16.8227 \\
    cross-att & 6.9067 & \underline{0.7222} & \underline{0.1791} & \underline{17.010} \\
    cross-self att & 6.8938 & \bf 0.7244 & \bf 0.1766 & \bf 17.1215 \\
    self-cross att & \bf \cc{6.7288} & \cc{0.7215} & \cc{0.1814} & \cc{16.9642} \\
    \end{tabular}}
    \end{center} }
    \end{minipage} }
    \hfill
    \subfloat[\textbf{Conditions Aggregation}. Combining pose feature $v_P$ with both local $v_L$ and global feature $v_G$ gives the best FID.
    \label{tab:aggregation}]{
    \centering
    \begin{minipage}{0.31\linewidth}{
    \begin{center}
    \resizebox{1.0\hsize}{!}{
    \begin{tabular}{ccccc}
    CANet & FID $\downarrow$ & SSIM $\uparrow$ & LPIPS $\downarrow$ & PSNR $\uparrow$ \\
    \midrule
    $v_L+v_P$ & 11.2661 & 0.6679 & 0.3131 & 12.8694 \\
    $v_P+v_G$ & 6.9241 & \bf 0.7230 & \bf 0.1805 & \bf 17.0456 \\
    $v_L+v_P+v_G$ & \bf \cc{6.7288} & \underline{\cc{0.7215}} & \underline{\cc{0.1814}} & \underline{\cc{16.9642}} \\
    \end{tabular}}
    \end{center} }
    \end{minipage} }
    \hfill
    \subfloat[\textbf{Inference speed}. It is averaged over 10 times for each 8 image generation with 50 DDIM steps ($256\times 176$), using one A100 GPU.
    \label{tab:inference_speed}]{
    \centering
    \begin{minipage}{0.34\linewidth}{
    \begin{center}
    \resizebox{1.0\hsize}{!}{
    \begin{tabular}{ccccc}
    Method & Param (M) $\downarrow$ & Time (s) $\downarrow$ & Speed up $\uparrow$ & FID $\downarrow$ \\
    \midrule
    PIDM & 688.00 & 16.975$\pm 0.055$ & $1.0\times$ & 6.36\\
    \bf X-MDPT-S & \bf 33.52 & \bf 1.191$\pm$0.021 & \bf 14.25$\times$ & 7.42 \\
    \bf X-MDPT-B & \bf 131.92 & \bf 1.299$\pm$0.022 & \bf 13.07$\times$ & 6.72 \\
    \bf X-MDPT-L & \bf 460.24 & \bf 3.124$\pm$0.026 & \bf 5.43$\times$ & \bf 6.25 \\
    \end{tabular} }
    \end{center} }
    \end{minipage} }
    \vspace{-12pt}
\end{table*}

\textbf{Inference Time.}
As presented in Tab. \ref{tab:inference_speed}, all variants of our models demonstrate significantly faster inference speeds and require fewer parameters compared to the runner-up, PIDM. Specifically, models X-MDPT-S, X-MDPT-B, and X-MDPT-L speed up PIDM by $14.25\times$, $13.07\times$, and $5.43\times$, respectively. This speed advantage can be attributed to several factors. Firstly, X-MDPT operates on latent patches of $32\times 32$, while PIDM works directly on pixel space of $256\times 256$. Here, the VAE in X-MDPT works a single forward, whereas PIDM necessitates 50 forward passes in DDIM. Secondly, PIDM employs disentangled classifier-free guidance for both pose and source, requiring two forwards and resulting in $50\times$ more evaluation steps. Conversely, our X-MDPT utilizes CFG for a unified condition and needs only one forward. The training time can be found in the \textbf{Appendix} due to space constraints, where our method proves significantly more efficient than PIDM.

\textbf{MIPNet Components.}
(a) \textbf{Mask Ratio}: In Tab. \ref{tab:masking}, we observe that a lower mask ratio of the target image in MIPNet yields the best FID score, consistent with findings from \citet{gao2023masked} on MDT applied to ImageNet. Conversely, a higher ratio, such as 70\%, results in better SSIM and LPIPS scores. This higher ratio compels the model to prioritize reconstruction over semantic representation learning, which is crucial for effective generation.

(b) \textbf{Attention for MIPNet}: Tab. \ref{tab:attention_ablation} illustrates that the self-cross attention design produces the best FID score while utilizing only self-attention (as MDT) yields the worst FID. This discrepancy highlights the significance of reference-based mask prediction in enhancing the performance of person image generation.

\textbf{CANet Components.}
(a) \textbf{Conditions Aggregation}: Tab. \ref{tab:aggregation} illustrates that solely utilizing the local feature $v_L$, \ie the VAE's output of the source image, yields the poorest performance (the $v_L+v_P$ case). Conversely, relying solely on the global feature $v_G$ (the $v_G+v_P$ case) leads to significant improvement. However, this approach lacks local information crucial for the generation, as the noisy target latent of TDNet operates at the VAE level. The optimal choice arises from combining both local and global features, as demonstrated in the $v_L+v_P+v_G$ case.

(b) \textbf{Global and Pose Representations}: AnyDoor \cite{chen2023anydoor} demonstrates that DINOv2-G is good at capturing object details. In Tab. \ref{tab:global} we observe that the quality of global features significantly impacts performance, with higher-capacity DINOv2 models yielding better FID scores.

For the pose, we find that DINOv2-B outperforms DINOv2-S in FID but slightly lags in SSIM and LPIPS. We opt for DINOv2-G for the global feature of $x_s$ and DINOv2-B for the pose $y_p$ as default. The simplicity of the pose image allows a smaller model like DINOv2-B to effectively guide TDNet. Conversely, the complexity of the human image necessitates a more powerful variant.

\section{More Discussions}
\label{discussions_sec}
During the discussion phase, we delved into additional properties of the proposed method, uncovering several findings.

\textbf{Compared with Prior Efficient Method:} PoCoLD introduced a latent diffusion more efficient than PIDM but falls behind FID. Since PoCoLD's code is not complete, we conducted our own implementation. Tab. \ref{tab:compare_memory} shows that our variants run faster PoCoLD on A100. Notably, X-MDPT-S, with $11\times$ fewer parameters, surpasses PoCoLD on FID, generating 8 images in just 1 second, $3\times$ faster than PoCoLD. 
\vspace{-8pt}
\begin{table}[!htbp]
    \caption{\textbf{Compare Efficiency.} Results are presented for every 8-image generation (batch size=8) for $256\times 176$ image, 50 denoising steps, using one NVIDIA A100 GPU. We conduct 10 runs and take the average.}
    \vspace{-10pt}
    \label{tab:compare_memory}
    \begin{center}
    \resizebox{1.0\hsize}{!}{
    \begin{tabular}{cccccc}
    \toprule
    \bf Method & \bf  Infer. Time (s) & \bf  Mem. Use (M) & \bf Param. (M) & \bf  Guidance & \bf  \#Forward \\
    \midrule
    PIDM & 16.975 $\pm$ 0.055 & 9572 & 688.00 & D-CFG & 150 \\
    PoCOLD & 3.215 $\pm$ 0.028 & 5686 & 395.89 &  D-CFG & 150 \\
    \midrule
    X-MDPT-L (Ours) & \bf 3.124 $\pm$ 0.026 & 6438 & 460.24 & CFG & \bf 100 \\
    X-MDPT-B (Ours) & \bf 1.299 $\pm$ 0.022 & \bf 5680 & \bf 131.92 &  CFG & \bf 100 \\
    X-MDPT-S (Ours) & \bf 1.191 $\pm$ 0.021 & \bf 5485 & \bf 33.52 &  CFG & \bf 100 \\
    \end{tabular}}
    \end{center}
\end{table}
\vspace{-12pt}
Our method's enhanced speed over PoCoLD arises from PoCoLD's use of cumulative CFG, a variant of Disentangled Classifier-Free Guidance (D-CFG, similar to PIDM), which significantly slows down its operation. Specifically, each generation in PoCoLD requires a total of 150 forwards. In 50 denoising steps, one step necessitates three forwards (unconditional, pose-condition, source-image condition), resulting in a total of $50 \times 3 = 150$.

In contrast, X-MDPT employs standard CFG, requiring only 100 forwards, where each denoising step needs two forwards (unconditional, unified condition), totaling $50 \times 2 = 100$, saving 50 forwards compared to PoCoLD and PIDM.

\textbf{Performance for Higher Resolution.}
In Tab. \ref{tab:compare_deepfashion}, for the DeepFashion dataset, X-MDPT shows a slightly lower FID score for higher-resolution images compared to lower-resolution ones, consistent with PIDM and NTED but not with PoCoLD. We suggest that our Transformer-based framework can better capture finer details and contextual information in higher-resolution images compared to PoCoLD, which relies on Unet. In contrast, on the ImageNet dataset, the behavior of Diffusion Transformer (DiT paper) is opposite to its performance on DeepFashion. We believe these disparities warrant a systematic exploration of dataset diversity and model architectures (CNN, Transformer, etc.) for a conclusive understanding.

\textbf{Insights of Adding MIPNet.}
We integrate MIPNet into the Transformer backbone during training to facilitate the model in leveraging contextual information among patches within an image (via self-attention) and learning the correspondence between the source and target image (via cross-attention). This enhances the Transformer's learning capacity more rapidly. This insight is corroborated by Tab. \ref{tab:all_ablations}(a), where MIPNet significantly improves performance compared to using the Transformer alone. The efficacy of masking loss, trained concurrently with the main loss, has been observed in previous works such as iBOT \cite{zhou2022image_ibot} and PatchMAE \cite{zhang2023patch} in the self-supervised representation learning domain.

\begin{figure}[!htbp]
  \centering
  \includegraphics[width=1.0\linewidth]{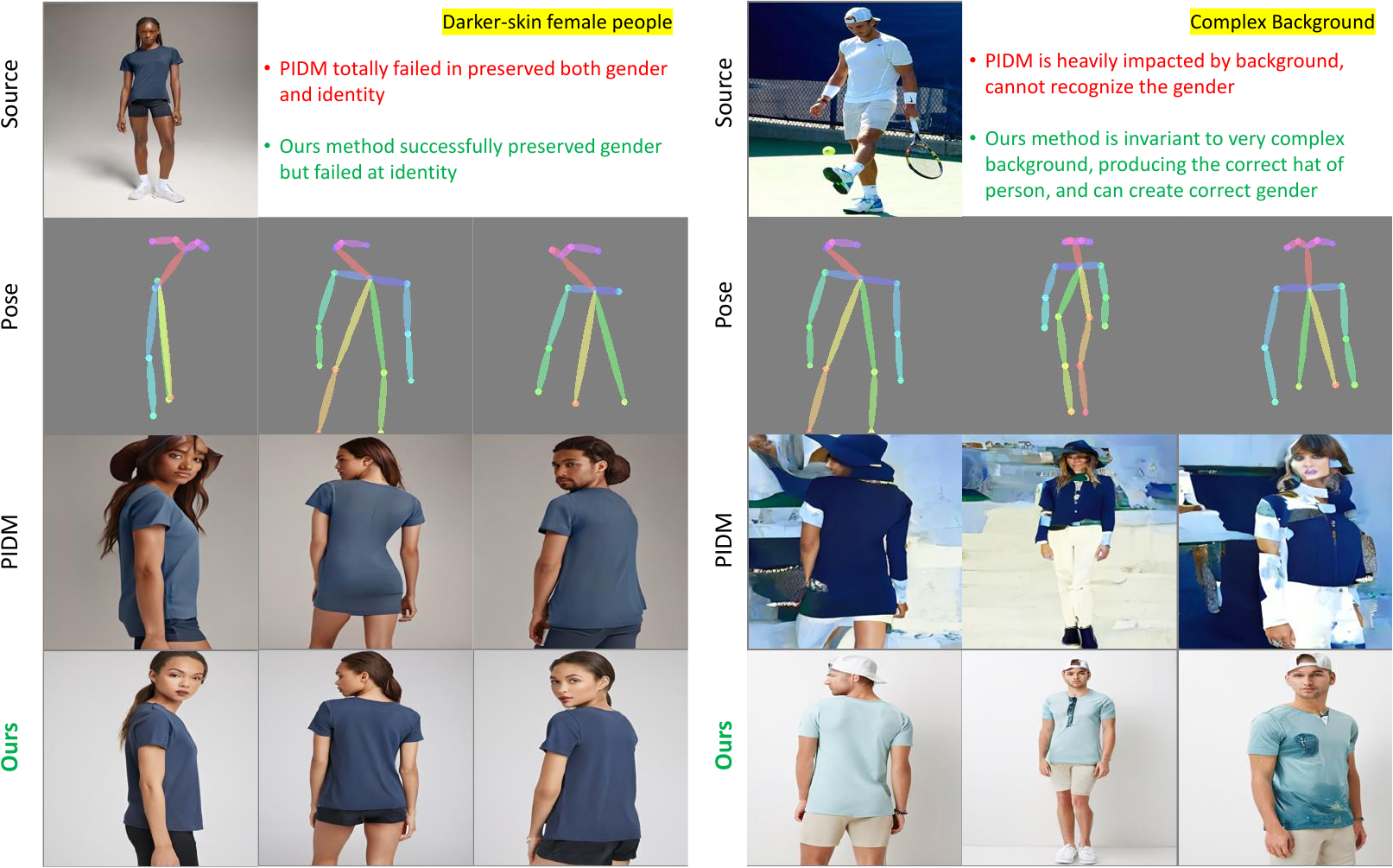}
  \vspace{-10pt}
  \caption{\textbf{In-the-wild testing.} Our X-MDPT-L consistently generates meaningful images with accurate gender representation, whereas PIDM \cite{bhunia2023person} fails.}
  \label{fig:in_the_wild_test}
  \vspace{-10pt}
\end{figure}

\textbf{Generalizations.} We have conducted an evaluation of in-the-wild images to prove the generalization of our method. We use our ready-to-use X-MDPT-L model trained on the DeepFashion dataset. We compare with the second-best method PIDM using their published checkpoint. We test with more challenging cases such as people with darker skin (as DeepFashion most data are collected with white people, a few for black) and complex background images. Fig. \ref{fig:in_the_wild_test} demonstrates that X-MDPT is much more stable than the runner-up method PIDM, highlighting the distinguishing properties of diffusion transformers on this task.
We believe one of the reasons PIDM suffers from overfitting more seriously is because it uses the features of pose and source image that are trained from scratch, while our method employs a pre-trained model DINO that gives much better features for both pose and source image. Training on more in-the-wild data can be a solution for mitigating the overfitting (toward background, identity) of ours and PIDM.

\textbf{Overlapping Cases.} Overlaps occur rarely, but in DeepFashion, we discovered several overlapping scenes, notably, a duplicated data pair with IDs \texttt{00005136} and \texttt{00005188}. Further details can be found in the Appendix. When testing on these duplicate samples, other approaches failed to reconstruct the target images. Why can X-MDPT fully reproduce previously seen samples while other models cannot? We attribute this capability to the Diffusion Transformer, which generates based on its training sample observations more effectively than previous CNN Unet-based algorithms.

\vspace{-8pt}
\section{Conclusion}
\vspace{-4pt}
\label{conclusions_sec}
In this paper, we present X-MDPT, a novel masked diffusion generative model for pose-guided human image generation (PHIG). Unlike previous methods using Unet for denoising diffusion, X-MDPT employs a Transformer on latent patches. Our analysis shows that X-MDPT achieves 99.99\% similarity in generating view-invariant vectors, ensuring consistent target images across poses. Extensive experiments demonstrate X-MDPT's efficiency in producing high-quality, high-resolution images, surpassing existing approaches in inference speed and setting a new state-of-the-art for PHIG on the DeepFashion benchmark.

\section*{Impact Statement}

Our method proficiently produces high-quality images featuring individuals in diverse poses, utilizing any person's image as a reference point. While it offers numerous advantages, including swift image generation, there's a risk of misuse, such as the creation of deceptive content for fraudulent purposes, a well-documented issue in image synthesis. We are committed to implementing measures to regulate access, thereby preventing misuse and ensuring that the technology contributes to the community's welfare safely.

\nocite{pham2023self}
\nocite{pham2022lad}
\nocite{pham2022pros}

\section*{Acknowledgements}
This work was supported by the Institute for Information \& communications Technology Planning \& Evaluation (IITP) grant funded by the Korea government (MSIT) (No. 2021-0-01381, \textit{Development of Causal AI through Video Understanding and Reinforcement Learning, and Its Applications to Real Environments}) and partly supported by IITP grant funded by the Korea government (MSIT) (No. 2022-0-00184, \textit{Development and Study of AI Technologies to Inexpensively Conform to Evolving Policy on Ethics}).

\bibliography{references}

\begin{thebibliography}{56}
\providecommand{\natexlab}[1]{#1}
\providecommand{\url}[1]{\texttt{#1}}
\expandafter\ifx\csname urlstyle\endcsname\relax
  \providecommand{\doi}[1]{doi: #1}\else
  \providecommand{\doi}{doi: \begingroup \urlstyle{rm}\Url}\fi

\bibitem[Bhunia et~al.(2023)Bhunia, Khan, Cholakkal, Anwer, Laaksonen, Shah,
  and Khan]{bhunia2023person}
Bhunia, A.~K., Khan, S., Cholakkal, H., Anwer, R.~M., Laaksonen, J., Shah, M.,
  and Khan, F.~S.
\newblock Person image synthesis via denoising diffusion model.
\newblock In \emph{Proceedings of the IEEE/CVF Conference on Computer Vision
  and Pattern Recognition}, pp.\  5968--5976, 2023.

\bibitem[Cao et~al.(2017)Cao, Simon, Wei, and Sheikh]{cao2017realtime}
Cao, Z., Simon, T., Wei, S.-E., and Sheikh, Y.
\newblock Realtime multi-person 2d pose estimation using part affinity fields.
\newblock In \emph{Proceedings of the IEEE conference on computer vision and
  pattern recognition}, pp.\  7291--7299, 2017.

\bibitem[Chang et~al.(2022)Chang, Zhang, Jiang, Liu, and
  Freeman]{chang2022maskgit}
Chang, H., Zhang, H., Jiang, L., Liu, C., and Freeman, W.~T.
\newblock Maskgit: Masked generative image transformer.
\newblock In \emph{Proceedings of the IEEE/CVF Conference on Computer Vision
  and Pattern Recognition}, pp.\  11315--11325, 2022.

\bibitem[Chang et~al.(2023)Chang, Zhang, Barber, Maschinot, Lezama, Jiang,
  Yang, Murphy, Freeman, Rubinstein, et~al.]{chang2023muse}
Chang, H., Zhang, H., Barber, J., Maschinot, A., Lezama, J., Jiang, L., Yang,
  M.-H., Murphy, K., Freeman, W.~T., Rubinstein, M., et~al.
\newblock Muse: Text-to-image generation via masked generative transformers.
\newblock \emph{arXiv preprint arXiv:2301.00704}, 2023.

\bibitem[Chen et~al.(2023)Chen, Huang, Liu, Shen, Zhao, and
  Zhao]{chen2023anydoor}
Chen, X., Huang, L., Liu, Y., Shen, Y., Zhao, D., and Zhao, H.
\newblock Anydoor: Zero-shot object-level image customization.
\newblock \emph{arXiv preprint arXiv:2307.09481}, 2023.

\bibitem[Devlin et~al.(2018)Devlin, Chang, Lee, and Toutanova]{devlin2018bert}
Devlin, J., Chang, M.-W., Lee, K., and Toutanova, K.
\newblock Bert: Pre-training of deep bidirectional transformers for language
  understanding.
\newblock \emph{arXiv preprint arXiv:1810.04805}, 2018.

\bibitem[Dosovitskiy et~al.(2010)Dosovitskiy, Beyer, Kolesnikov, Weissenborn,
  Zhai, Unterthiner, Dehghani, Minderer, Heigold, Gelly,
  et~al.]{dosovitskiy2010image}
Dosovitskiy, A., Beyer, L., Kolesnikov, A., Weissenborn, D., Zhai, X.,
  Unterthiner, T., Dehghani, M., Minderer, M., Heigold, G., Gelly, S., et~al.
\newblock An image is worth 16x16 words: Transformers for image recognition at
  scale. arxiv 2020.
\newblock \emph{arXiv preprint arXiv:2010.11929}, 2010.

\bibitem[Dosovitskiy et~al.(2021)Dosovitskiy, Beyer, Kolesnikov, Weissenborn,
  Zhai, Unterthiner, Dehghani, Minderer, Heigold, Gelly, Uszkoreit, and
  Houlsby]{dosovitskiy2021an}
Dosovitskiy, A., Beyer, L., Kolesnikov, A., Weissenborn, D., Zhai, X.,
  Unterthiner, T., Dehghani, M., Minderer, M., Heigold, G., Gelly, S.,
  Uszkoreit, J., and Houlsby, N.
\newblock An image is worth 16x16 words: Transformers for image recognition at
  scale.
\newblock In \emph{ICLR}, 2021.

\bibitem[Gao et~al.(2023)Gao, Zhou, Cheng, and Yan]{gao2023masked}
Gao, S., Zhou, P., Cheng, M.-M., and Yan, S.
\newblock Masked diffusion transformer is a strong image synthesizer.
\newblock \emph{arXiv preprint arXiv:2303.14389}, 2023.

\bibitem[Gupta et~al.(2023)Gupta, Wu, Deng, and Fei-Fei]{gupta2023siamese}
Gupta, A., Wu, J., Deng, J., and Fei-Fei, L.
\newblock Siamese masked autoencoders.
\newblock In \emph{Thirty-seventh Conference on Neural Information Processing
  Systems}, 2023.
\newblock URL \url{https://openreview.net/forum?id=yC3q7vInux}.

\bibitem[Han et~al.(2023)Han, Zhu, Deng, Song, and Xiang]{han2023controllable}
Han, X., Zhu, X., Deng, J., Song, Y.-Z., and Xiang, T.
\newblock Controllable person image synthesis with pose-constrained latent
  diffusion.
\newblock In \emph{Proceedings of the IEEE/CVF International Conference on
  Computer Vision}, pp.\  22768--22777, 2023.

\bibitem[He et~al.(2022)He, Chen, Xie, Li, Doll{\'a}r, and
  Girshick]{he2022masked}
He, K., Chen, X., Xie, S., Li, Y., Doll{\'a}r, P., and Girshick, R.
\newblock Masked autoencoders are scalable vision learners.
\newblock In \emph{Proceedings of the IEEE/CVF conference on computer vision
  and pattern recognition}, pp.\  16000--16009, 2022.

\bibitem[Ho \& Salimans(2022)Ho and Salimans]{ho2022classifier}
Ho, J. and Salimans, T.
\newblock Classifier-free diffusion guidance.
\newblock \emph{arXiv preprint arXiv:2207.12598}, 2022.

\bibitem[Ho et~al.(2020)Ho, Jain, and Abbeel]{ho2020denoising}
Ho, J., Jain, A., and Abbeel, P.
\newblock Denoising diffusion probabilistic models.
\newblock \emph{Advances in neural information processing systems},
  33:\penalty0 6840--6851, 2020.

\bibitem[Ho et~al.(2022)Ho, Saharia, Chan, Fleet, Norouzi, and
  Salimans]{ho2022cascaded}
Ho, J., Saharia, C., Chan, W., Fleet, D.~J., Norouzi, M., and Salimans, T.
\newblock Cascaded diffusion models for high fidelity image generation.
\newblock \emph{The Journal of Machine Learning Research}, 23\penalty0
  (1):\penalty0 2249--2281, 2022.

\bibitem[Isola et~al.(2017)Isola, Zhu, Zhou, and Efros]{isola2017image}
Isola, P., Zhu, J.-Y., Zhou, T., and Efros, A.~A.
\newblock Image-to-image translation with conditional adversarial networks.
\newblock In \emph{Proceedings of the IEEE conference on computer vision and
  pattern recognition}, pp.\  1125--1134, 2017.

\bibitem[Jung et~al.(2020)Jung, Hong, Wang, Han, Pham, Park, Kim, Kang, Yoo,
  and Lee]{jung2020flexible}
Jung, Y.~H., Hong, S.~K., Wang, H.~S., Han, J.~H., Pham, T.~X., Park, H., Kim,
  J., Kang, S., Yoo, C.~D., and Lee, K.~J.
\newblock Flexible piezoelectric acoustic sensors and machine learning for
  speech processing.
\newblock \emph{Advanced Materials}, 32\penalty0 (35):\penalty0 1904020, 2020.

\bibitem[Jung et~al.(2022)Jung, Pham, Issa, Wang, Lee, Chung, Lee, Kim, Yoo,
  and Lee]{jung2022deep}
Jung, Y.~H., Pham, T.~X., Issa, D., Wang, H.~S., Lee, J.~H., Chung, M., Lee,
  B.-Y., Kim, G., Yoo, C.~D., and Lee, K.~J.
\newblock Deep learning-based noise robust flexible piezoelectric acoustic
  sensors for speech processing.
\newblock \emph{Nano Energy}, 101:\penalty0 107610, 2022.

\bibitem[Karras et~al.(2023)Karras, Holynski, Wang, and
  Kemelmacher-Shlizerman]{karras2023dreampose}
Karras, J., Holynski, A., Wang, T.-C., and Kemelmacher-Shlizerman, I.
\newblock Dreampose: Fashion image-to-video synthesis via stable diffusion.
\newblock \emph{arXiv preprint arXiv:2304.06025}, 2023.

\bibitem[Kim et~al.(2020)Kim, Ma, Pham, Kim, and Yoo]{kim2020modality}
Kim, J., Ma, M., Pham, T., Kim, K., and Yoo, C.~D.
\newblock Modality shifting attention network for multi-modal video question
  answering.
\newblock In \emph{Proceedings of the IEEE/CVF conference on computer vision
  and pattern recognition}, pp.\  10106--10115, 2020.

\bibitem[Lee et~al.(2020)Lee, Park, Pham, and Yoo]{lee2020learning}
Lee, D., Park, H., Pham, T., and Yoo, C.~D.
\newblock Learning augmentation network via influence functions.
\newblock In \emph{Proceedings of the IEEE/CVF Conference on Computer Vision
  and Pattern Recognition}, pp.\  10961--10970, 2020.

\bibitem[Liu et~al.(2016)Liu, Luo, Qiu, Wang, and Tang]{liu2016deepfashion}
Liu, Z., Luo, P., Qiu, S., Wang, X., and Tang, X.
\newblock Deepfashion: Powering robust clothes recognition and retrieval with
  rich annotations.
\newblock In \emph{Proceedings of the IEEE conference on computer vision and
  pattern recognition}, pp.\  1096--1104, 2016.

\bibitem[Ma et~al.(2017)Ma, Jia, Sun, Schiele, Tuytelaars, and
  Van~Gool]{ma2017pose}
Ma, L., Jia, X., Sun, Q., Schiele, B., Tuytelaars, T., and Van~Gool, L.
\newblock Pose guided person image generation.
\newblock \emph{Advances in neural information processing systems}, 30, 2017.

\bibitem[Men et~al.(2020)Men, Mao, Jiang, Ma, and Lian]{men2020controllable}
Men, Y., Mao, Y., Jiang, Y., Ma, W.-Y., and Lian, Z.
\newblock Controllable person image synthesis with attribute-decomposed gan.
\newblock In \emph{Proceedings of the IEEE/CVF conference on computer vision
  and pattern recognition}, pp.\  5084--5093, 2020.

\bibitem[Niu et~al.(2023)Niu, Zhang, Pham, Sun, Zhu, Kweon, and
  Zhang]{niu2023cdpmsr}
Niu, A., Zhang, K., Pham, T.~X., Sun, J., Zhu, Y., Kweon, I.~S., and Zhang, Y.
\newblock Cdpmsr: Conditional diffusion probabilistic models for single image
  super-resolution.
\newblock In \emph{2023 IEEE International Conference on Image Processing
  (ICIP)}, pp.\  615--619. IEEE, 2023.

\bibitem[Niu et~al.(2024{\natexlab{a}})Niu, Pham, Zhang, Sun, Zhu, Yan, Kweon,
  and Zhang]{niu2024acdmsr}
Niu, A., Pham, T.~X., Zhang, K., Sun, J., Zhu, Y., Yan, Q., Kweon, I.~S., and
  Zhang, Y.
\newblock Acdmsr: Accelerated conditional diffusion models for single image
  super-resolution.
\newblock \emph{IEEE Transactions on Broadcasting}, 2024{\natexlab{a}}.

\bibitem[Niu et~al.(2024{\natexlab{b}})Niu, Zhang, Pham, Wang, Sun, Kweon, and
  Zhang]{niu2024learning}
Niu, A., Zhang, K., Pham, T.~X., Wang, P., Sun, J., Kweon, I.~S., and Zhang, Y.
\newblock Learning from multi-perception features for real-word image
  super-resolution.
\newblock \emph{IEEE Transactions on Circuits and Systems for Video
  Technology}, 2024{\natexlab{b}}.

\bibitem[Oquab et~al.(2023)Oquab, Darcet, Moutakanni, Vo, Szafraniec, Khalidov,
  Fernandez, Haziza, Massa, El-Nouby, Howes, Huang, Xu, Sharma, Li, Galuba,
  Rabbat, Assran, Ballas, Synnaeve, Misra, Jegou, Mairal, Labatut, Joulin, and
  Bojanowski]{oquab2023dinov2}
Oquab, M., Darcet, T., Moutakanni, T., Vo, H.~V., Szafraniec, M., Khalidov, V.,
  Fernandez, P., Haziza, D., Massa, F., El-Nouby, A., Howes, R., Huang, P.-Y.,
  Xu, H., Sharma, V., Li, S.-W., Galuba, W., Rabbat, M., Assran, M., Ballas,
  N., Synnaeve, G., Misra, I., Jegou, H., Mairal, J., Labatut, P., Joulin, A.,
  and Bojanowski, P.
\newblock Dinov2: Learning robust visual features without supervision, 2023.

\bibitem[Peebles \& Xie(2023)Peebles and Xie]{peebles2023scalable}
Peebles, W. and Xie, S.
\newblock Scalable diffusion models with transformers.
\newblock In \emph{Proceedings of the IEEE/CVF International Conference on
  Computer Vision}, pp.\  4195--4205, 2023.

\bibitem[Pham et~al.(2022{\natexlab{a}})Pham, Zhang, Niu, Zhang, and
  Yoo]{pham2022pros}
Pham, T., Zhang, C., Niu, A., Zhang, K., and Yoo, C.~D.
\newblock On the pros and cons of momentum encoder in self-supervised visual
  representation learning.
\newblock \emph{arXiv preprint arXiv:2208.05744}, 2022{\natexlab{a}}.

\bibitem[Pham et~al.(2021)Pham, Mina, Issa, and Yoo]{pham2021self}
Pham, T.~X., Mina, R. J.~L., Issa, D., and Yoo, C.~D.
\newblock Self-supervised learning with local attention-aware feature.
\newblock \emph{arXiv preprint arXiv:2108.00475}, 2021.

\bibitem[Pham et~al.(2022{\natexlab{b}})Pham, Mina, Nguyen, Madjid, Choi, and
  Yoo]{pham2022lad}
Pham, T.~X., Mina, R. J.~L., Nguyen, T., Madjid, S.~R., Choi, J., and Yoo,
  C.~D.
\newblock Lad: A hybrid deep learning system for benign paroxysmal positional
  vertigo disorders diagnostic.
\newblock \emph{IEEE Access}, 2022{\natexlab{b}}.

\bibitem[Pham et~al.(2023)Pham, Niu, Zhang, Jin, Hong, and Yoo]{pham2023self}
Pham, T.~X., Niu, A., Zhang, K., Jin, T. J.~T., Hong, J.~W., and Yoo, C.~D.
\newblock Self-supervised visual representation learning via residual momentum.
\newblock \emph{IEEE Access}, 2023.

\bibitem[Ren et~al.(2020)Ren, Yu, Chen, Li, and Li]{ren2020deep}
Ren, Y., Yu, X., Chen, J., Li, T.~H., and Li, G.
\newblock Deep image spatial transformation for person image generation.
\newblock In \emph{Proceedings of the IEEE/CVF Conference on Computer Vision
  and Pattern Recognition}, pp.\  7690--7699, 2020.

\bibitem[Ren et~al.(2022)Ren, Fan, Li, Liu, and Li]{ren2022neural}
Ren, Y., Fan, X., Li, G., Liu, S., and Li, T.~H.
\newblock Neural texture extraction and distribution for controllable person
  image synthesis.
\newblock In \emph{Proceedings of the IEEE/CVF Conference on Computer Vision
  and Pattern Recognition}, pp.\  13535--13544, 2022.

\bibitem[Rombach et~al.(2022)Rombach, Blattmann, Lorenz, Esser, and
  Ommer]{rombach2022high}
Rombach, R., Blattmann, A., Lorenz, D., Esser, P., and Ommer, B.
\newblock High-resolution image synthesis with latent diffusion models.
\newblock In \emph{Proceedings of the IEEE/CVF conference on computer vision
  and pattern recognition}, pp.\  10684--10695, 2022.

\bibitem[Ronneberger et~al.(2015)Ronneberger, Fischer, and
  Brox]{ronneberger2015u}
Ronneberger, O., Fischer, P., and Brox, T.
\newblock U-net: Convolutional networks for biomedical image segmentation.
\newblock In \emph{International Conference on Medical image computing and
  computer-assisted intervention}, 2015.

\bibitem[Siarohin et~al.(2018)Siarohin, Sangineto, Lathuiliere, and
  Sebe]{siarohin2018deformable}
Siarohin, A., Sangineto, E., Lathuiliere, S., and Sebe, N.
\newblock Deformable gans for pose-based human image generation.
\newblock In \emph{Proceedings of the IEEE conference on computer vision and
  pattern recognition}, pp.\  3408--3416, 2018.

\bibitem[Song et~al.(2020)Song, Meng, and Ermon]{song2020denoising}
Song, J., Meng, C., and Ermon, S.
\newblock Denoising diffusion implicit models.
\newblock \emph{arXiv preprint arXiv:2010.02502}, 2020.

\bibitem[Song \& Ermon(2019)Song and Ermon]{song2019generative}
Song, Y. and Ermon, S.
\newblock Generative modeling by estimating gradients of the data distribution.
\newblock \emph{Advances in neural information processing systems}, 32, 2019.

\bibitem[Trung \& Yoo(2019)Trung and Yoo]{trungshort}
Trung, P.~X. and Yoo, C.~D.
\newblock Short convolutional neural network and mfccs for accurate speaker
  recognition systems.
\newblock \emph{International Technical Conference on Circuits/Systems,
  Computers and Communications (ITC-CSCC)}, 2019.

\bibitem[Vaswani et~al.(2017)Vaswani, Shazeer, Parmar, Uszkoreit, Jones, Gomez,
  Kaiser, and Polosukhin]{vaswani2017attention}
Vaswani, A., Shazeer, N., Parmar, N., Uszkoreit, J., Jones, L., Gomez, A.~N.,
  Kaiser, L., and Polosukhin, I.
\newblock Attention is all you need.
\newblock In \emph{NeurIPS}, 2017.

\bibitem[Vu et~al.(2019)Vu, Jang, Pham, and Yoo]{vu2019cascade}
Vu, T., Jang, H., Pham, T.~X., and Yoo, C.
\newblock Cascade rpn: Delving into high-quality region proposal network with
  adaptive convolution.
\newblock \emph{Advances in neural information processing systems}, 32, 2019.

\bibitem[Wu et~al.(2023)Wu, Si, Wang, Qu, and Jing]{wu2023pose}
Wu, J., Si, S., Wang, J., Qu, X., and Jing, X.
\newblock Pose guided human image synthesis with partially decoupled gan.
\newblock In \emph{Asian Conference on Machine Learning}, pp.\  1133--1148.
  PMLR, 2023.

\bibitem[Yu et~al.(2023)Yu, Cheng, Sohn, Lezama, Zhang, Chang, Hauptmann, Yang,
  Hao, Essa, et~al.]{yu2023magvit}
Yu, L., Cheng, Y., Sohn, K., Lezama, J., Zhang, H., Chang, H., Hauptmann,
  A.~G., Yang, M.-H., Hao, Y., Essa, I., et~al.
\newblock Magvit: Masked generative video transformer.
\newblock In \emph{Proceedings of the IEEE/CVF Conference on Computer Vision
  and Pattern Recognition}, pp.\  10459--10469, 2023.

\bibitem[Zhang et~al.(2022{\natexlab{a}})Zhang, Zhang, Pham, Niu, Qiao, Yoo,
  and Kweon]{zhang2022dual}
Zhang, C., Zhang, K., Pham, T.~X., Niu, A., Qiao, Z., Yoo, C.~D., and Kweon,
  I.~S.
\newblock Dual temperature helps contrastive learning without many negative
  samples: Towards understanding and simplifying moco.
\newblock In \emph{Proceedings of the IEEE/CVF Conference on Computer Vision
  and Pattern Recognition}, pp.\  14441--14450, 2022{\natexlab{a}}.

\bibitem[Zhang et~al.(2022{\natexlab{b}})Zhang, Zhang, Zhang, Pham, Yoo, and
  Kweon]{zhang2022how}
Zhang, C., Zhang, K., Zhang, C., Pham, T.~X., Yoo, C.~D., and Kweon, I.~S.
\newblock How does simsiam avoid collapse without negative samples? a unified
  understanding with self-supervised contrastive learning.
\newblock In \emph{International Conference on Learning Representations},
  2022{\natexlab{b}}.
\newblock URL \url{https://openreview.net/forum?id=bwq6O4Cwdl}.

\bibitem[Zhang et~al.(2021)Zhang, Li, Lai, and Yang]{zhang2021pise}
Zhang, J., Li, K., Lai, Y.-K., and Yang, J.
\newblock Pise: Person image synthesis and editing with decoupled gan.
\newblock In \emph{Proceedings of the IEEE/CVF Conference on Computer Vision
  and Pattern Recognition}, pp.\  7982--7990, 2021.

\bibitem[Zhang et~al.(2023{\natexlab{a}})Zhang, Rao, and
  Agrawala]{zhang2023adding}
Zhang, L., Rao, A., and Agrawala, M.
\newblock Adding conditional control to text-to-image diffusion models,
  2023{\natexlab{a}}.

\bibitem[Zhang et~al.(2022{\natexlab{c}})Zhang, Yang, Lai, and
  Xie]{zhang2022exploring}
Zhang, P., Yang, L., Lai, J.-H., and Xie, X.
\newblock Exploring dual-task correlation for pose guided person image
  generation.
\newblock In \emph{Proceedings of the IEEE/CVF Conference on Computer Vision
  and Pattern Recognition}, pp.\  7713--7722, 2022{\natexlab{c}}.

\bibitem[Zhang et~al.(2023{\natexlab{b}})Zhang, Zhou, Wang, Wang, and
  Yan]{zhang2023patch}
Zhang, S., Zhou, Q., Wang, Z., Wang, F., and Yan, J.
\newblock Patch-level contrastive learning via positional query for visual
  pre-training.
\newblock In \emph{International Conference on Machine Learning}, pp.\
  41990--41999. PMLR, 2023{\natexlab{b}}.

\bibitem[Zhao et~al.(2023)Zhao, Chen, Chen, Bao, Hao, Yuan, and
  Wong]{zhao2023uni}
Zhao, S., Chen, D., Chen, Y.-C., Bao, J., Hao, S., Yuan, L., and Wong, K.-Y.~K.
\newblock Uni-controlnet: All-in-one control to text-to-image diffusion models.
\newblock \emph{Advances in Neural Information Processing Systems}, 2023.

\bibitem[Zhou et~al.(2022{\natexlab{a}})Zhou, Wei, Wang, Shen, Xie, Yuille, and
  Kong]{zhou2022image_ibot}
Zhou, J., Wei, C., Wang, H., Shen, W., Xie, C., Yuille, A., and Kong, T.
\newblock Image {BERT} pre-training with online tokenizer.
\newblock In \emph{International Conference on Learning Representations},
  2022{\natexlab{a}}.
\newblock URL \url{https://openreview.net/forum?id=ydopy-e6Dg}.

\bibitem[Zhou et~al.(2021)Zhou, Zhang, Zhang, Zhang, Bao, Chen, Zhang, and
  Wen]{zhou2021cocosnet}
Zhou, X., Zhang, B., Zhang, T., Zhang, P., Bao, J., Chen, D., Zhang, Z., and
  Wen, F.
\newblock Cocosnet v2: Full-resolution correspondence learning for image
  translation.
\newblock In \emph{Proceedings of the IEEE/CVF Conference on Computer Vision
  and Pattern Recognition}, pp.\  11465--11475, 2021.

\bibitem[Zhou et~al.(2022{\natexlab{b}})Zhou, Yin, Chen, Sun, Gao, and
  Li]{zhou2022cross}
Zhou, X., Yin, M., Chen, X., Sun, L., Gao, C., and Li, Q.
\newblock Cross attention based style distribution for controllable person
  image synthesis.
\newblock In \emph{European Conference on Computer Vision}, pp.\  161--178.
  Springer, 2022{\natexlab{b}}.

\bibitem[Zhu et~al.(2019)Zhu, Huang, Shi, Yu, Wang, and
  Bai]{zhu2019progressive}
Zhu, Z., Huang, T., Shi, B., Yu, M., Wang, B., and Bai, X.
\newblock Progressive pose attention transfer for person image generation.
\newblock In \emph{Proceedings of the IEEE/CVF Conference on Computer Vision
  and Pattern Recognition}, pp.\  2347--2356, 2019.

\end{thebibliography}
\bibliographystyle{icml2024}

\newpage
\appendix
\onecolumn



\section{Appendix}
\label{appendix_sec}

\subsection{Compare with the previous best-reported images}
\begin{figure*}[!htbp]
  \centering
  \includegraphics[width=1.0\linewidth]{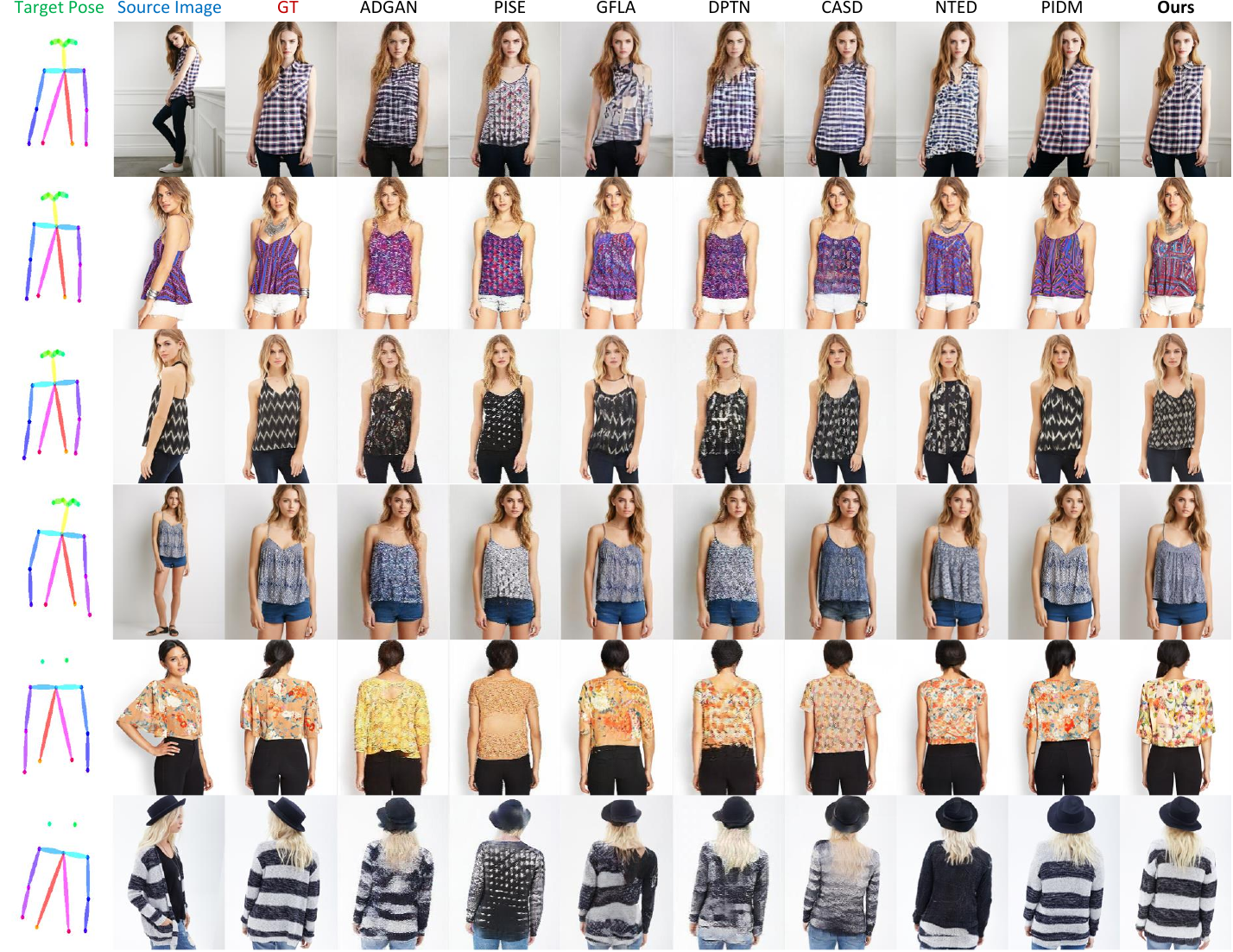}
  \vspace{-14pt}
  \caption{Compare some reported images in the prior paper and \textbf{ours}. Images generated by our method are also comparable or even better as it is closer to the ground truth images (\eg samples in rows 1, 2, 6). Images at $256\times 256$ resolution are resized to $256\times 176$.}
  \label{fig:single7}
\end{figure*}

To be more complete, in Fig. \ref{fig:single7} we also compare the best-generated images reported in the prior papers for reference. Our model consistently produces comparable or better outputs. 

\vspace{-18pt}

\vspace{-8pt}
\subsection{Generation given the same pose and different source views and vice versa }
\begin{figure*}[!htbp]
  \centering
  \includegraphics[width=1.0\linewidth]{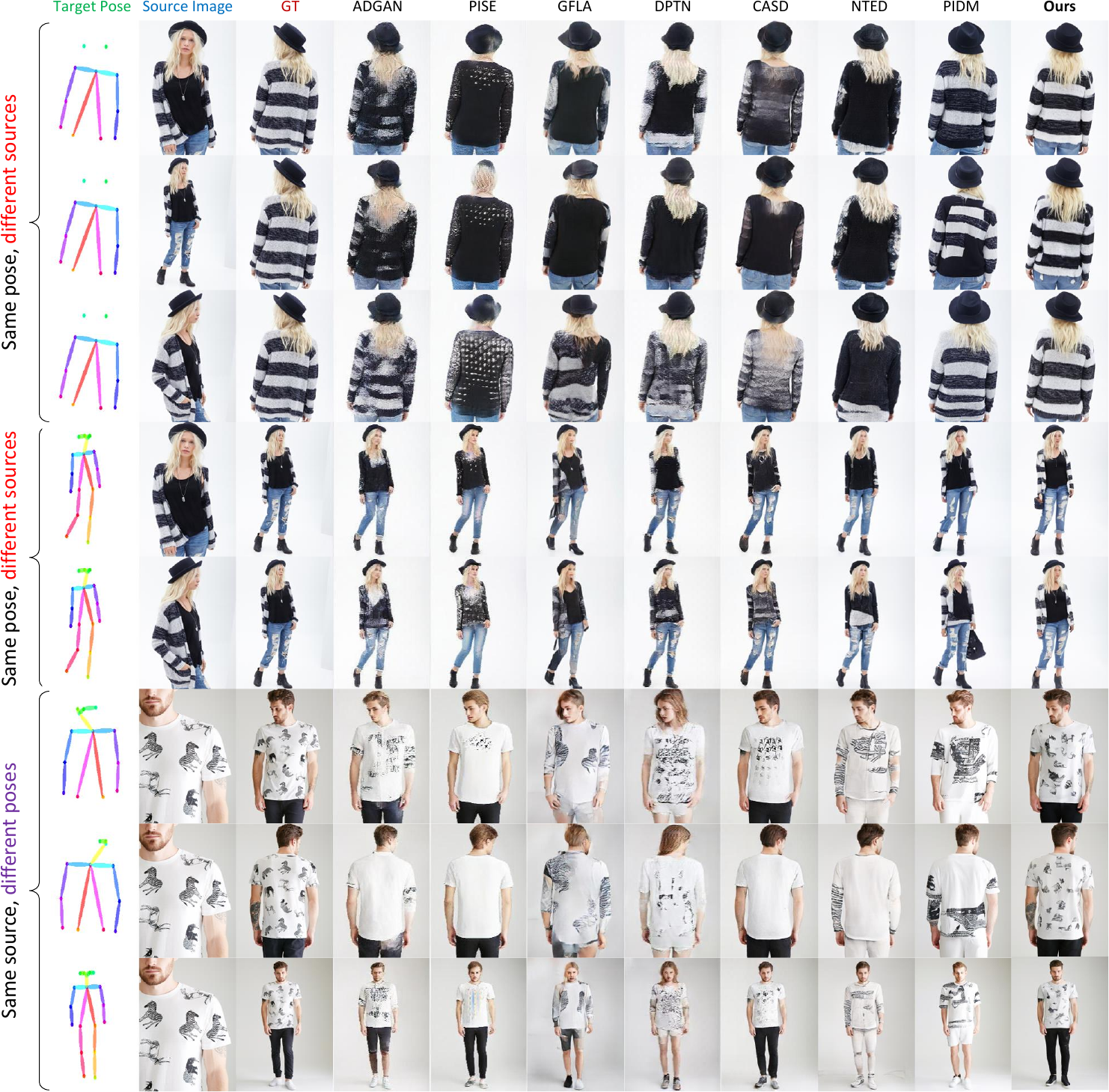}
  \vspace{-6pt}
  \caption{ \textbf{Different views to the same target pose and vice versa}. X-MDPT generates a target image that is more comprehensive and closely aligns with the ground truth image. Images at $256\times 256$ resolution are resized to $256\times 176$. }
  \label{fig:single8}
\end{figure*}

\begin{figure*}[!htbp]
  \centering
  \includegraphics[width=1.0\linewidth]{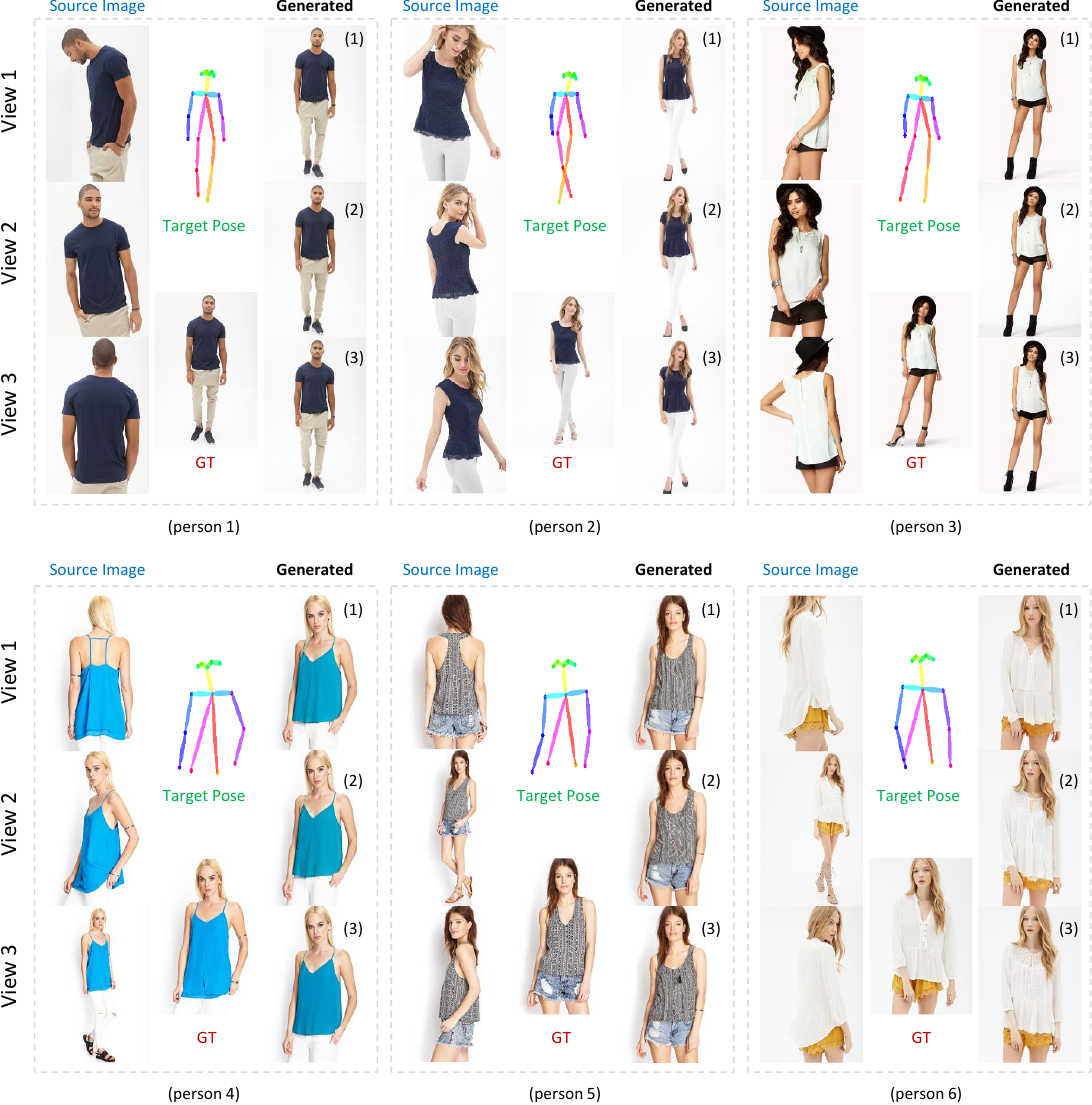}
  \vspace{-6pt}
  \caption{ \textbf{Different views of different persons}. X-MDPT-L generates consistent target images for the same target pose with three corresponding views, \textcolor{red}{GT} denotes ground truth. Images at $256\times 256$ resolution are resized to $256\times 176$. }
  \label{fig:view_invariant_appendix}
\end{figure*}

\vspace{-8pt}
In Fig. \ref{fig:single8}, we provide more comprehensive examples for comparison when all state-of-the-art approaches perform the generation with the given the same target pose + different views of the source image of the same person and vice versa. As shown in Fig. \ref{fig:single8}, in the case of given the same target pose, when varying the source views of the person, all existing methods failed to capture the consistent target image, while our proposed X-MDPT handles it very well. Our model's superiority is also demonstrated when we keep the source image unchanged and generate different target images. Fig. \ref{fig:view_invariant_appendix} shows more generated by our X-MDPT-L for different persons with different views when generating the same target pose.

\vspace{-10pt}

\subsection{More details of experimental setups}
\begin{figure}[!htbp]
  \centering
  \includegraphics[width=0.6\linewidth]{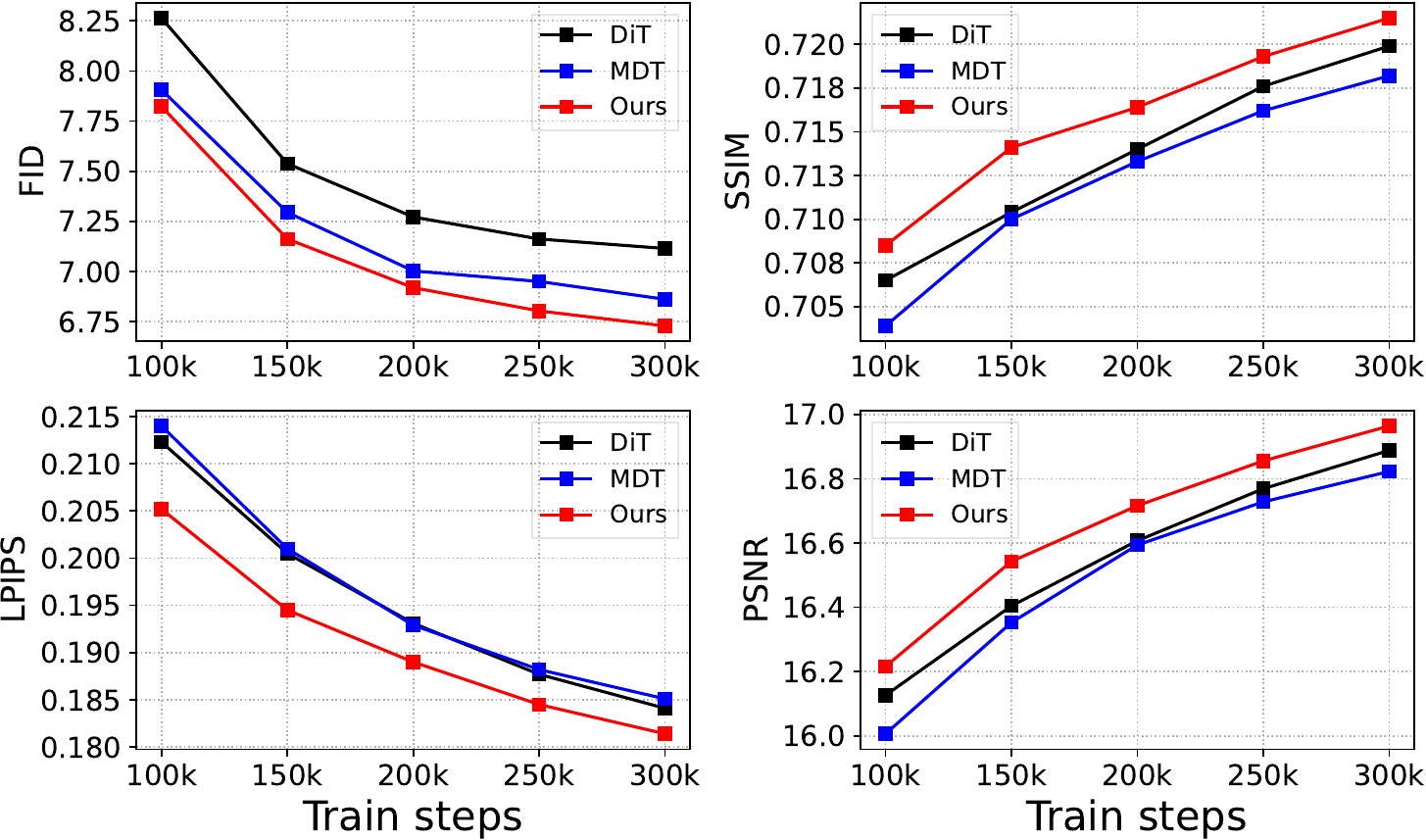}
  \caption{ \textbf{Transformer baselines.} We compare three frameworks: Applying to the PHIG problem with 1) DiT, 2) MDT, and 3) Ours. All methods use the size ViT-B (131M). MDT can improve DiT on FID but NOT on other metrics, while Ours improves on both metrics.
  }
  \label{fig:ablation_baselines}
  \vspace{-16pt}
\end{figure}

\vspace{-4pt}
We use 50 steps DDIM \cite{song2020denoising} for inference which is the same as PIDM. The details for three variants of X-MDPT-S, B, and L are provided in Tab. \ref{tab:configuration} and Fig.\ref{fig:ablation_baselines}. For VAE, we fine-tuned only the decoder using VAE of Stable Diffusion \cite{rombach2022high} on the training data of DeepFashion for 77 epochs. The face is distorted if not fine-tuning.

\vspace{-10pt}
\begin{table}[!htbp]
    \caption{\textbf{Parameters and Configs}. We follow ViT \cite{dosovitskiy2010image} to name models for Small (S), Base (B), and Large (L). Our X-MDPT has slightly more parameters compared to DiT \cite{peebles2023scalable} and MDT \cite{gao2023masked} as it needs one more cross-attention block, but overall, the total parameters are almost similar. }
    \label{tab:configuration}
    \begin{center}
    \resizebox{1\hsize}{!}{
    \begin{tabular}{ccccc|ccccc|ccccc}
    \toprule
    Method & Layers & Dim. & Heads & Param. (M) & Method & Layers & Dim. & Heads & Param. (M) & Method & Layers & Dim. & Heads & Param. (M) \\
    \midrule
    DiT-S & 12 & 384 & 6 & 32.9 & MDT-S & 12 & 384 & 6 & 33.1 & X-MDPT-S & 12 & 384 & 6 & 33.52 \\
    DiT-B & 12 & 768 & 12 & 130.3 & MDT-B & 12 & 768 & 12 & 130.8 & X-MDPT-B & 12 & 768 & 12 & 131.92 \\
    DiT-L & 24 & 1024 & 16 & 458.0 & MDT-L & 24 & 1024 & 16 & 459.1 & X-MDPT-L & 24 & 1024 & 16 & 460.24 \\
    \bottomrule
    \end{tabular}}
    \end{center}
\end{table}

\subsection{Failure cases}
\begin{figure}[!htbp]
  \centering
  \includegraphics[width=0.8\linewidth]{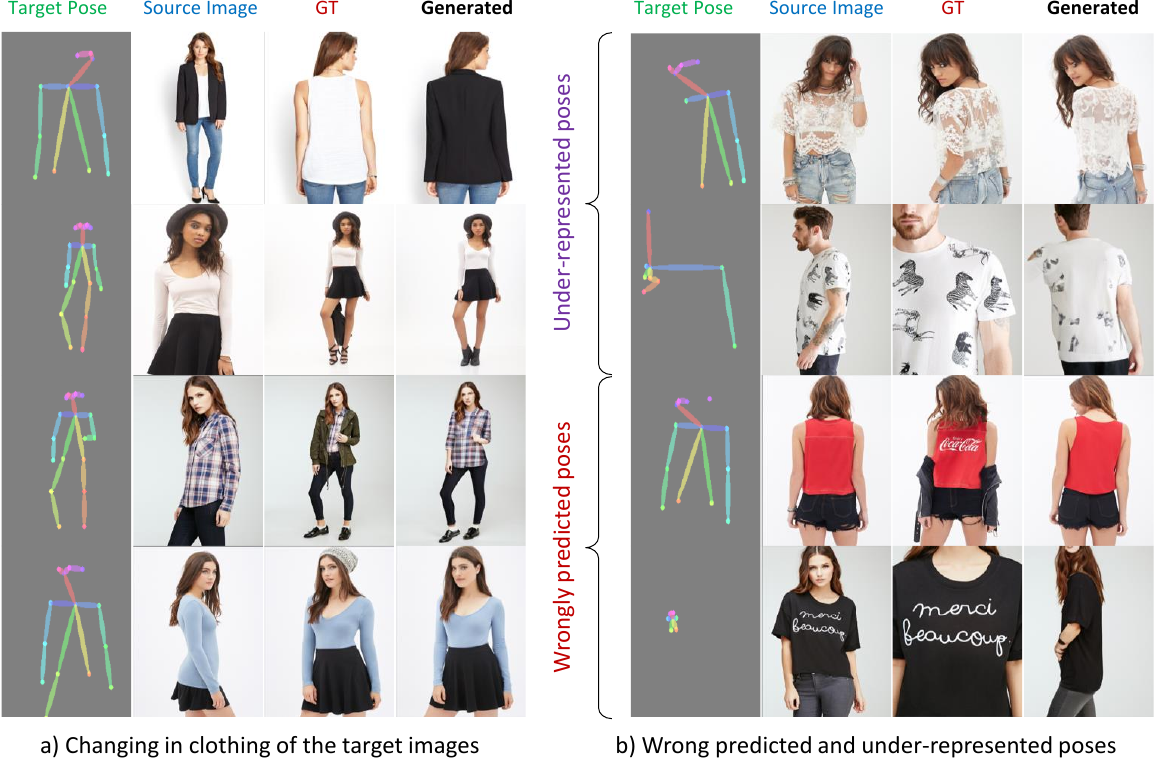}
  \caption{ \textbf{Failure cases.} Several use cases can result in worse metrics such as SSIM and LPIPS when compared to GT.
  }
  \label{fig:failure_cases}
\end{figure}

We note that our methods failed in certain cases that potentially mitigate it to achieve the best performance on the test set with SSIM, and LPIPS, as shown in Fig. \ref{fig:failure_cases}. First, changing in clothing of the target images can lead to worse quantitative scores. Second, the pose is under-represented, for example, the left hand of the woman is missing in the pose image making the model predict without that part. Third, the wrong-predicted pose that came from the OpenPose \cite{cao2017realtime} can make the model predict wrong targets. Increasing more training pairs of clothes changing and improving detected pose algorithms may resolve these failures.
\vspace{-10pt}
\vspace{-16pt}

\subsection{Different random seeds}
\begin{figure*}[!htbp]
  \centering
  \includegraphics[width=0.97\linewidth]{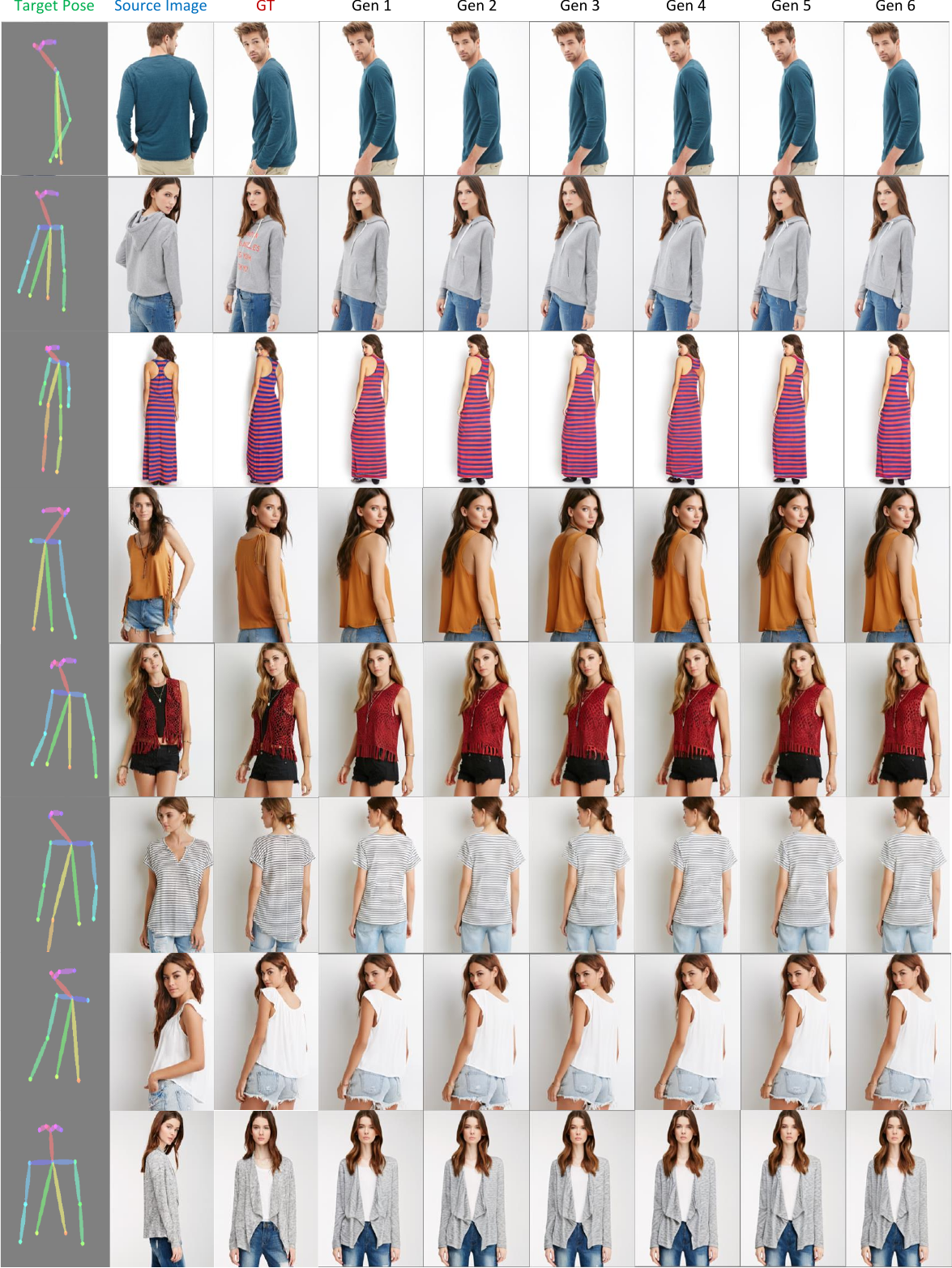}
  \vspace{-6pt}
  \caption{ \textbf{Different random seeds}. X-MDPT-L generates stable samples for 6 different random seeds, \textcolor{red}{GT} denotes ground truth. Images at $256\times 256$ resolution are resized to $256\times 176$. }
  \label{fig:seeds}
\end{figure*}

In Fig. \ref{fig:seeds} we show some samples generated by X-MDPT-L for six different seeds: 0, 100, 200, 300, 400, 500. As shown in the figure, our method generated the target images quite stable, demonstrating that it is not sensitive to the random seeds.

\subsection{Training time}
We compare the training times of our method and the runner-up PIDM \cite{bhunia2023person}. We refer to DreamPose \cite{karras2023dreampose} showed that PIDM uses 4 A100 GPUs trained for 26 days (section 2.3 of Dreampose), which results in if the use of the same resource as ours, \ie only 1 GPU A100, it would expect to need 104 days for PIDM. While our method trains for 800k steps (with a single GPU) takes 15.7 days. However, training our method using a single GPU with only 300k steps (5.9 days for X-MDPT-L, and 4.1 days for X-MDPT-B) already produces very good-quality images, while PIDM (needs 2/3 of their training to get a model can create good enough image, \ie would take approximately 69.3 days).
\begin{table}[!htbp]
    \caption{\textbf{Training time}. Compare PIDM and our method at $256\times 256$ resolution. The results show that our method is much more efficient in training time compared to the pixel-based PIDM when using the same computation resource (1 GPU). The inference time taken from our main paper is for generating 8 images using a single A100 GPU.}
    \label{tab:training_time}
    \begin{center}
    \resizebox{0.8\hsize}{!}{
    \begin{tabular}{cccc}
    \toprule
    Method & Training time to full converged $\downarrow$ & Inference Time (s) $\downarrow$ & Param. (M)$\downarrow$ \\
    \midrule
    PIDM \cite{bhunia2023person} & 104 days & 16.975 & 688.0 \\
    \bf X-MDPT-L & \bf 15 days & \bf 3.124 & \bf 460.24 \\
    \bottomrule
    \end{tabular}}
    \end{center}
\end{table}

We have not taken PoCoLD \cite{han2023controllable} to compare qualitative results because we find that the published code of PoCoLD is not complete, no published checkpoint, and we are unable to reproduce it. Instead, we show some references excerpted from their paper as follows (Table 2 in PoCoLD paper). Inference time of PIDM 9.25s vs. PoCoLD 4.99s where they measured on $256\times 256$ generation for a single GPU Tesla V100. We can see that PoCoLD only speeds up $1.85\times$ over PIDM, but our method speeds up PIDM with $14.25\times$, $13.07\times$, and $5.43\times$ for three of our model variants in the same settings. Note that, our smallest model X-MDPT-S already outperforms PoCoLD on FID score with $11\times$ fewer parameters and its inference speed is $14.25\times$ faster than PIDM.

We notice that the same PIDM, PoCoLD also uses the disentangled classifier free guidance form, so it will need one forward for unconditional, one forward for the pose condition, and one for the source image condition and results in 50 times more forwards, this will significantly mitigate their inference speed. By contrast, our method used normal CFG that required only one forward for condition and one for unconditional, saving 50 times of forward.

\subsection{Self-Supervised Learning Models}
There are various SSL models have been explored to learn the representations without labels \cite{he2022masked, pham2021self, pham2023self, oquab2023dinov2, zhang2022dual, zhang2022how}. These models serve as a good extractor for various applications \cite{pham2022lad, chen2023anydoor}. DINOv2 \cite{oquab2023dinov2} demonstrated an excellent pre-trained model for various diffusion-based frameworks. We mainly use DINOv2, but the other options may be worth trying. With the potential of diffusion transformers for conditional learning, it is expected to have more discovery of its capability in various domains and applications such as speech processing \cite{jung2022deep, jung2020flexible, trungshort}, data augmentation \cite{lee2020learning}, VQA \cite{kim2020modality}, visual detection learning \cite{vu2019cascade}, super-resolution \cite{niu2023cdpmsr,niu2024acdmsr,niu2024learning}.

\subsection{Overlapping in DeepFashion Dataset}
\begin{figure*}[!htbp]
  \centering
  \includegraphics[width=1.0\linewidth]{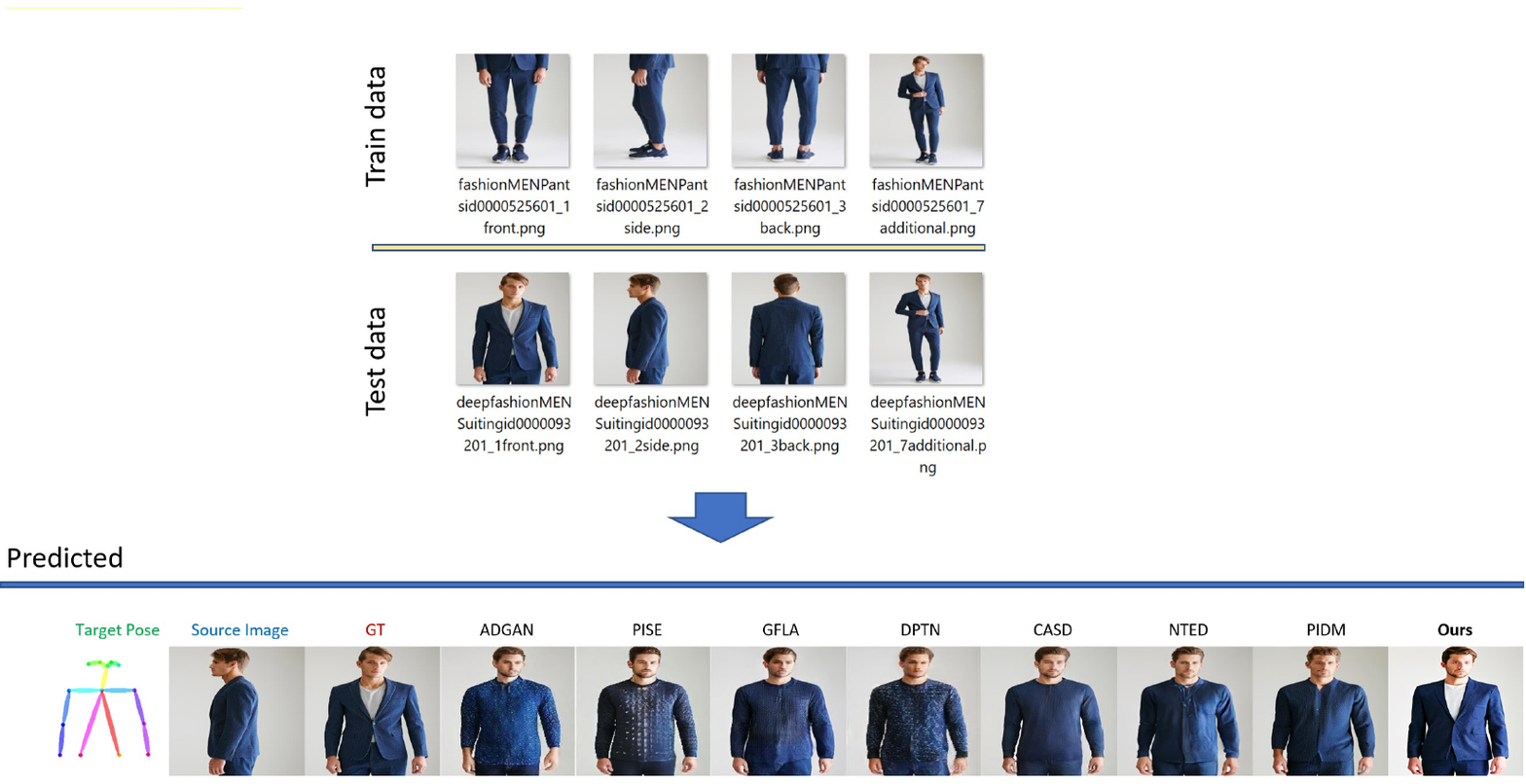}
  \vspace{-6pt}
  \caption{ Overlapping case on the DeepFashion dataset (1). }
  \label{fig:overlapping_1}
\end{figure*}

\begin{figure*}[!htbp]
  \centering
  \includegraphics[width=1.0\linewidth]{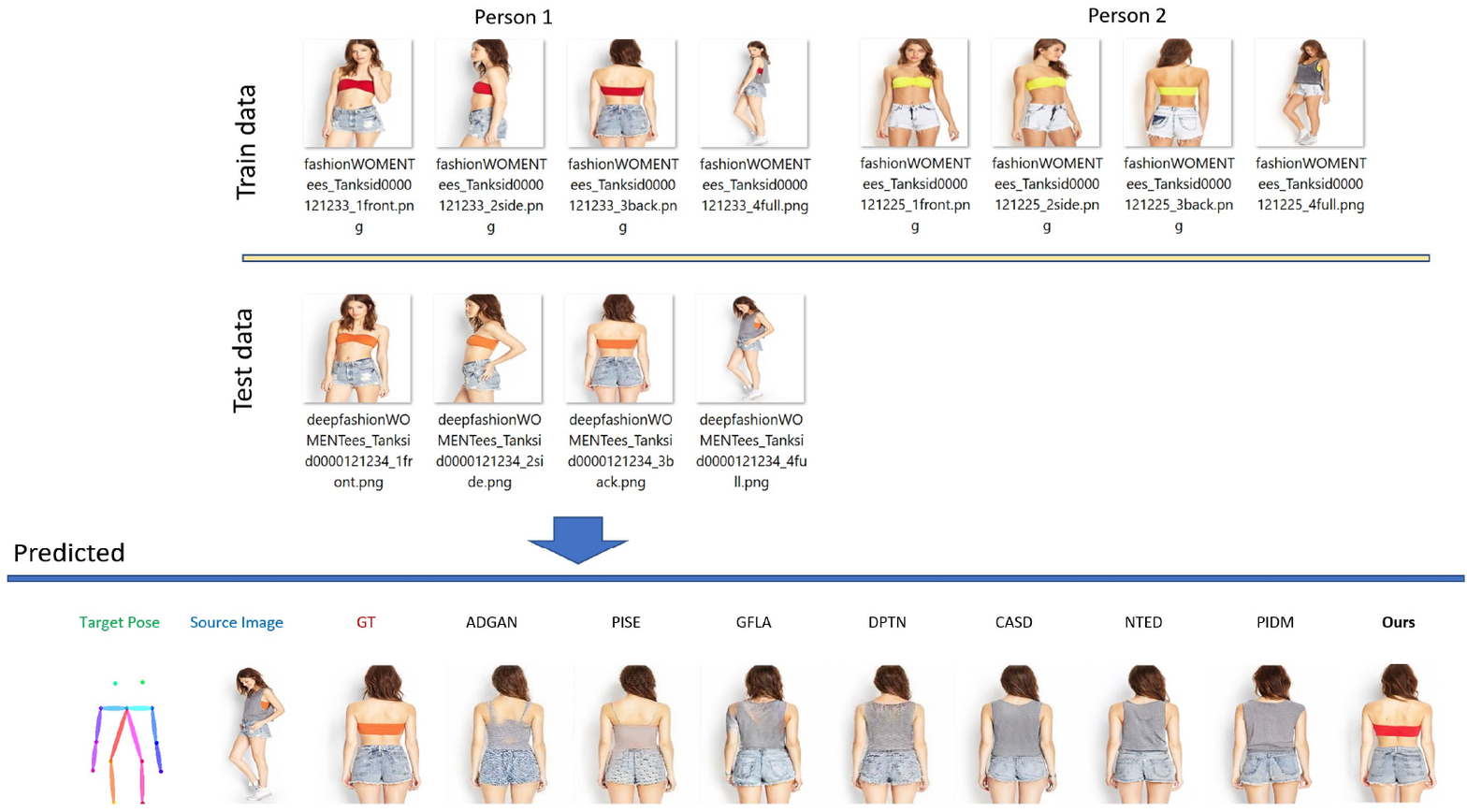}
  \vspace{-6pt}
  \caption{ Overlapping case on the DeepFashion dataset (2). }
  \label{fig:overlapping_2}
\end{figure*}

\begin{figure*}[!htbp]
  \centering
  \includegraphics[width=1.0\linewidth]{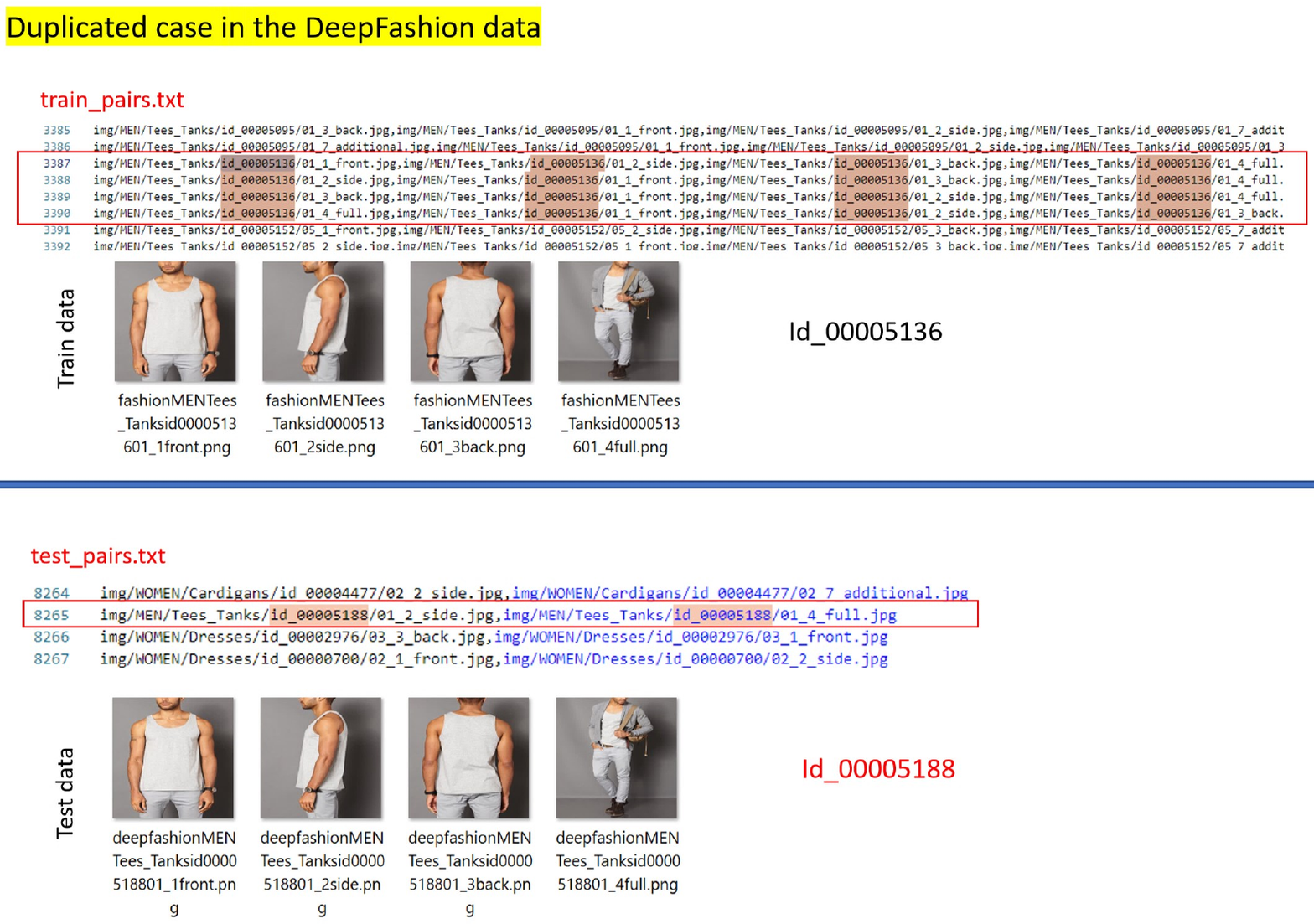}
  \vspace{-6pt}
  \caption{ Duplicated case on the DeepFashion dataset. }
  \label{fig:duplicated}
\end{figure*}

We illustrate several cases of overlapping in Fig. \ref{fig:overlapping_1} and Fig. \ref{fig:overlapping_2}, with a rare duplication case shown in Fig. \ref{fig:duplicated}.

\subsection{High-resolution images $512\times 512$}
\begin{figure*}[!htbp]
  \centering
  \includegraphics[width=1.0\linewidth]{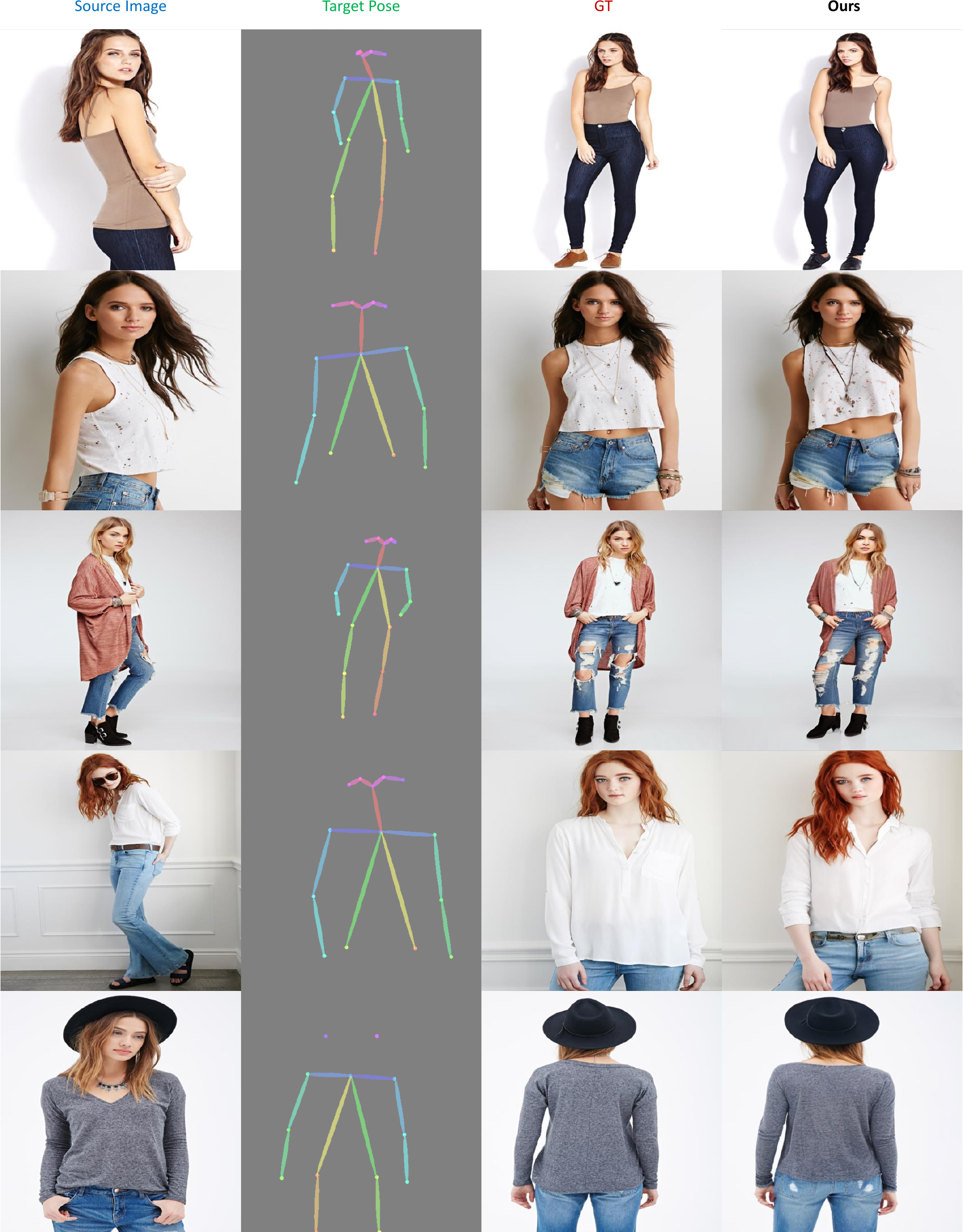}
  \vspace{-6pt}
  \caption{ $512\times 512$ Images generated by our X-MDPT-L model training on the DeepFashion dataset (1). }
  \label{fig:512x512}
\end{figure*}

\begin{figure*}[!htbp]
  \centering
  \includegraphics[width=1.0\linewidth]{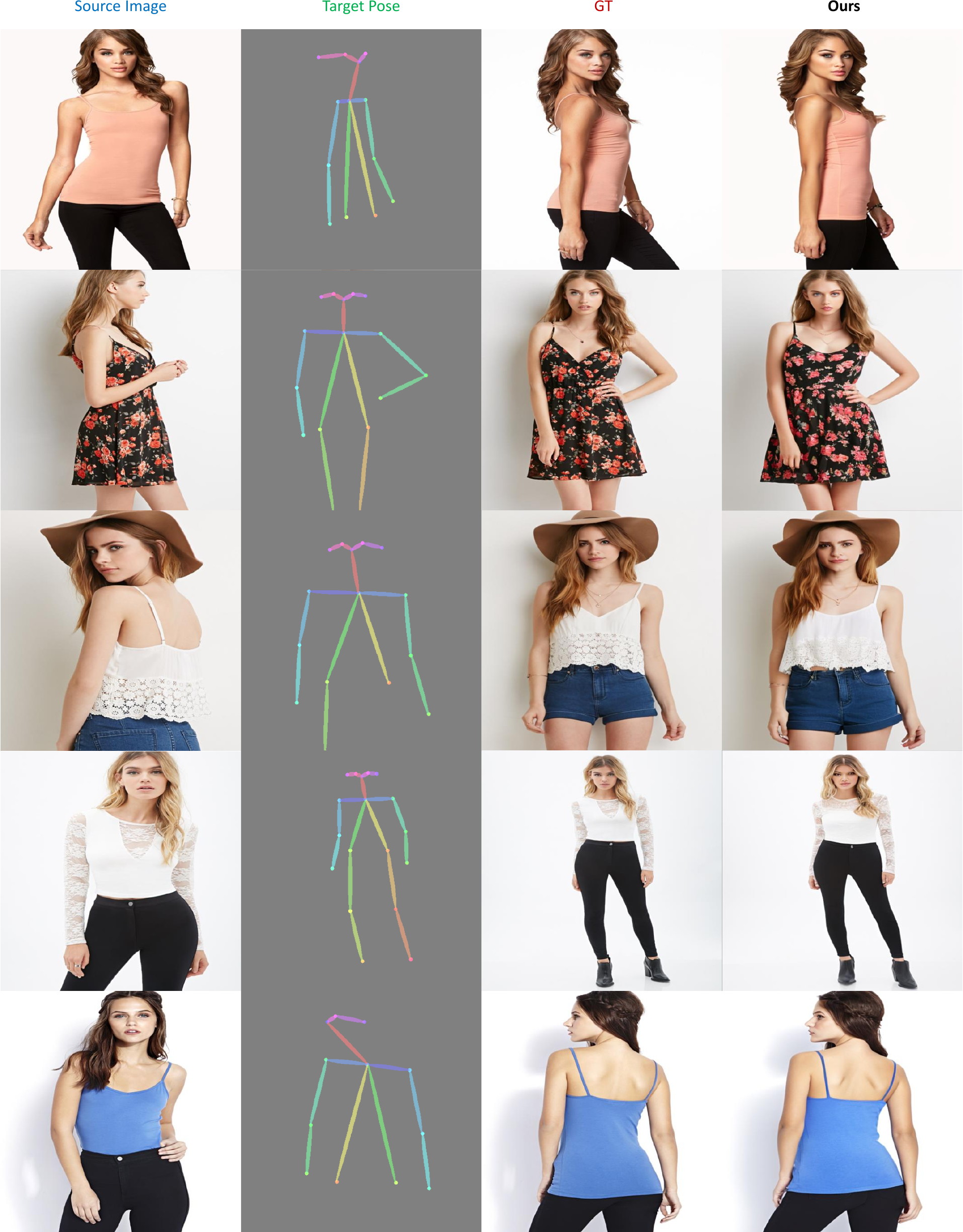}
  \vspace{-6pt}
  \caption{ $512\times 512$ Images generated by our X-MDPT-L model training on the DeepFashion dataset (2). }
  \label{fig:512x512_2}
\end{figure*}

\begin{figure*}[!htbp]
  \centering
  \includegraphics[width=1.0\linewidth]{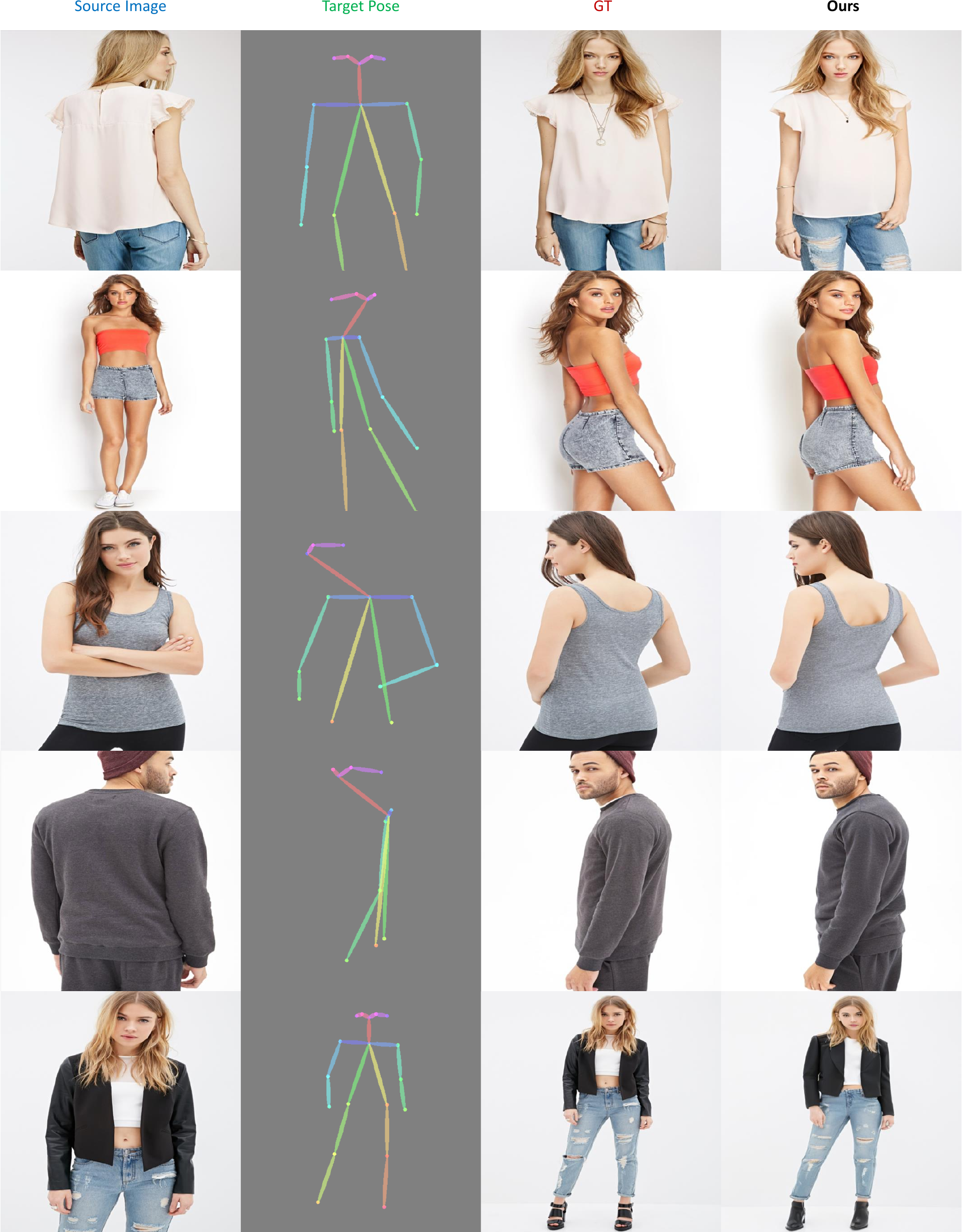}
  \vspace{-6pt}
  \caption{ $512\times 512$ Images generated by our X-MDPT-L model training on the DeepFashion dataset (3). }
  \label{fig:512x512_3}
\end{figure*}

We also report some high-resolution images $512\times 512$ generated by our method in Fig. \ref{fig:512x512}, Fig. \ref{fig:512x512_2}, and Fig. \ref{fig:512x512_3}, etc... Model X-MDPT-L is trained on the DeepFashion dataset for high resolution and generates excellent target images.

\begin{figure*}[!htbp]
  \centering
  \includegraphics[width=1.0\linewidth]{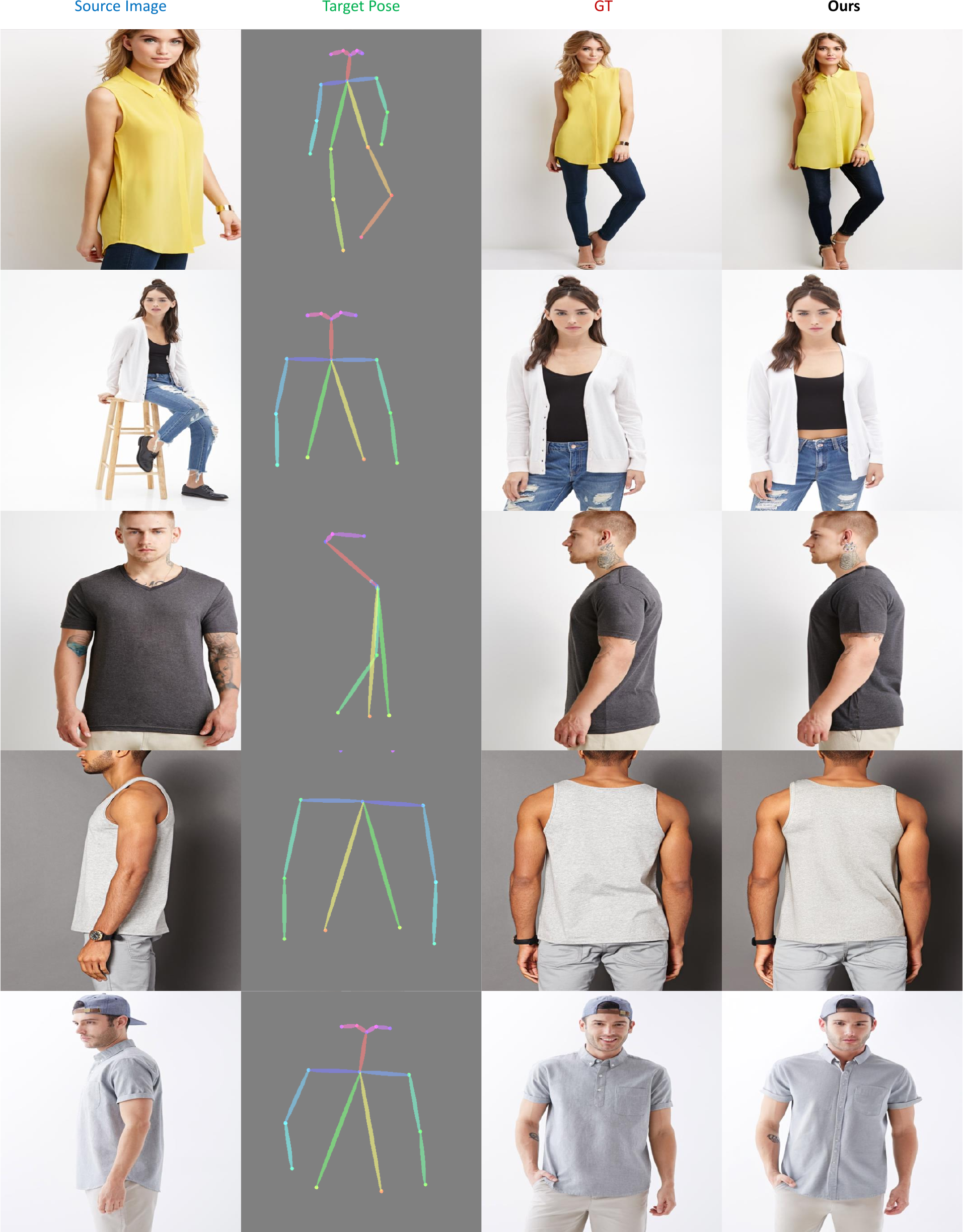}
  \vspace{-6pt}
  \caption{ $512\times 512$ Images generated by our X-MDPT-L model training on the DeepFashion dataset (4). }
  \label{fig:512x512_4}
\end{figure*}

\begin{figure*}[!htbp]
  \centering
  \includegraphics[width=1.0\linewidth]{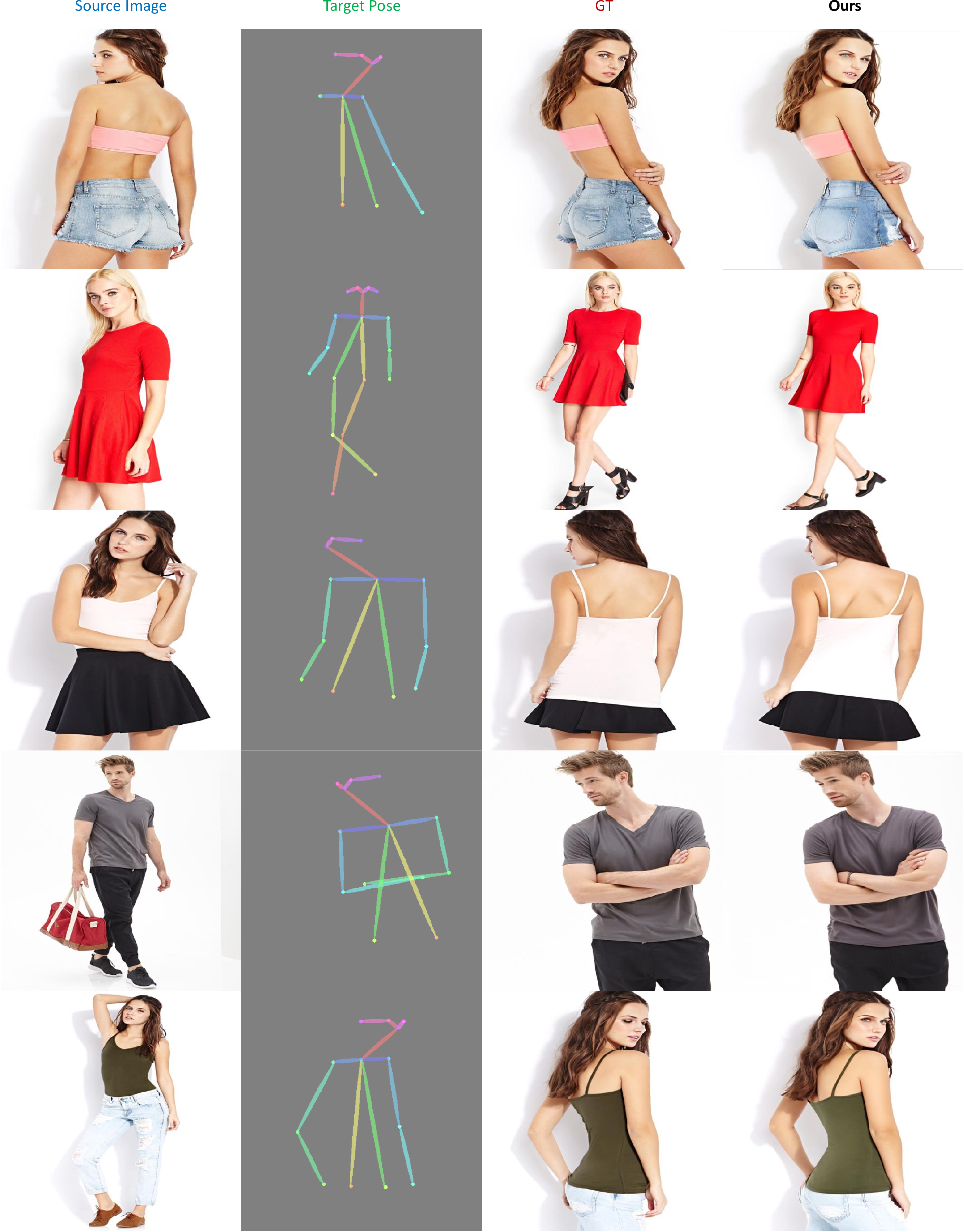}
  \vspace{-6pt}
  \caption{ $512\times 512$ Images generated by our X-MDPT-L model training on the DeepFashion dataset (5). }
  \label{fig:512x512_5}
\end{figure*}

\newpage
\subsection{More qualitative comparisons on $256\times 176$}
More visualization results in resolution $256\times 176$ are provided below from Fig. \ref{fig:single1} below. For various difficult cases such as rare poses and a close look at the source image, other methods failed to generate the correct target, while X-MDPT can handle them adequately.

\begin{figure*}[!htbp]
  \centering
  \includegraphics[width=1.0\linewidth]{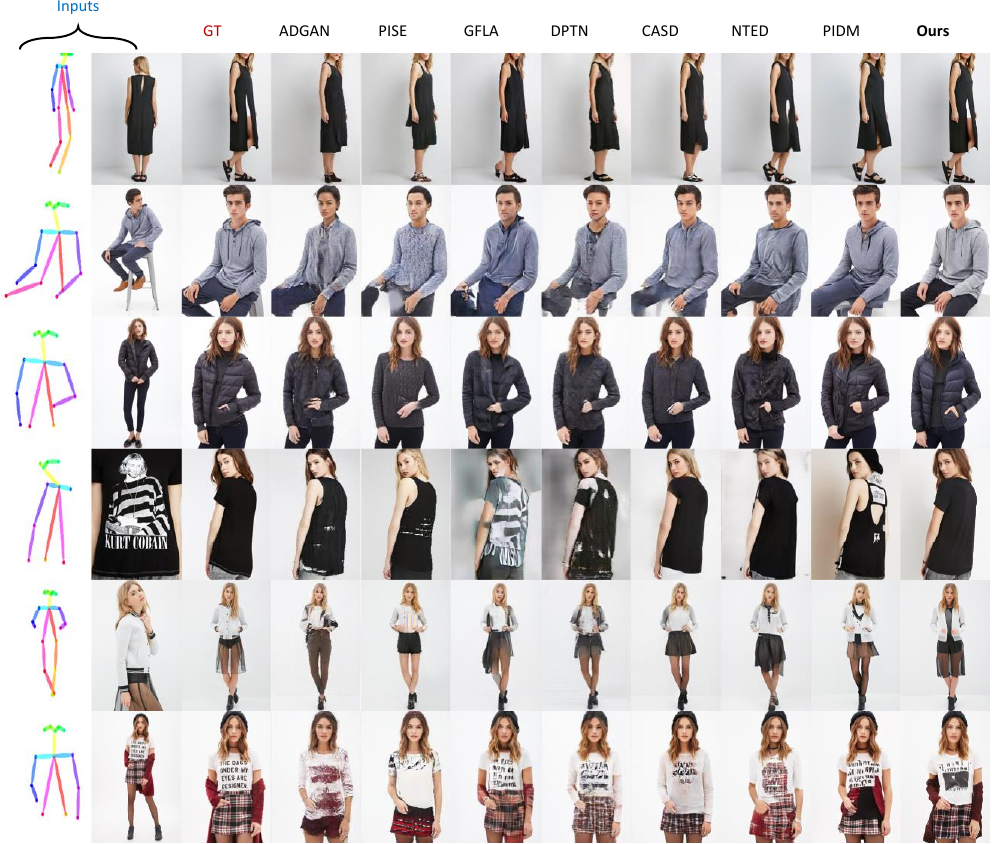}
  \vspace{-6pt}
  \caption{ More comparison images with state-of-the-art approaches. X-MDPT generates a target image that is more comprehensive and closely aligns with the ground truth image. Images at $256\times 256$ resolution are resized to $256\times 176$. (1) }
  \label{fig:single1}
\end{figure*}

\begin{figure*}[!htbp]
  \centering
  \includegraphics[width=1.0\linewidth]{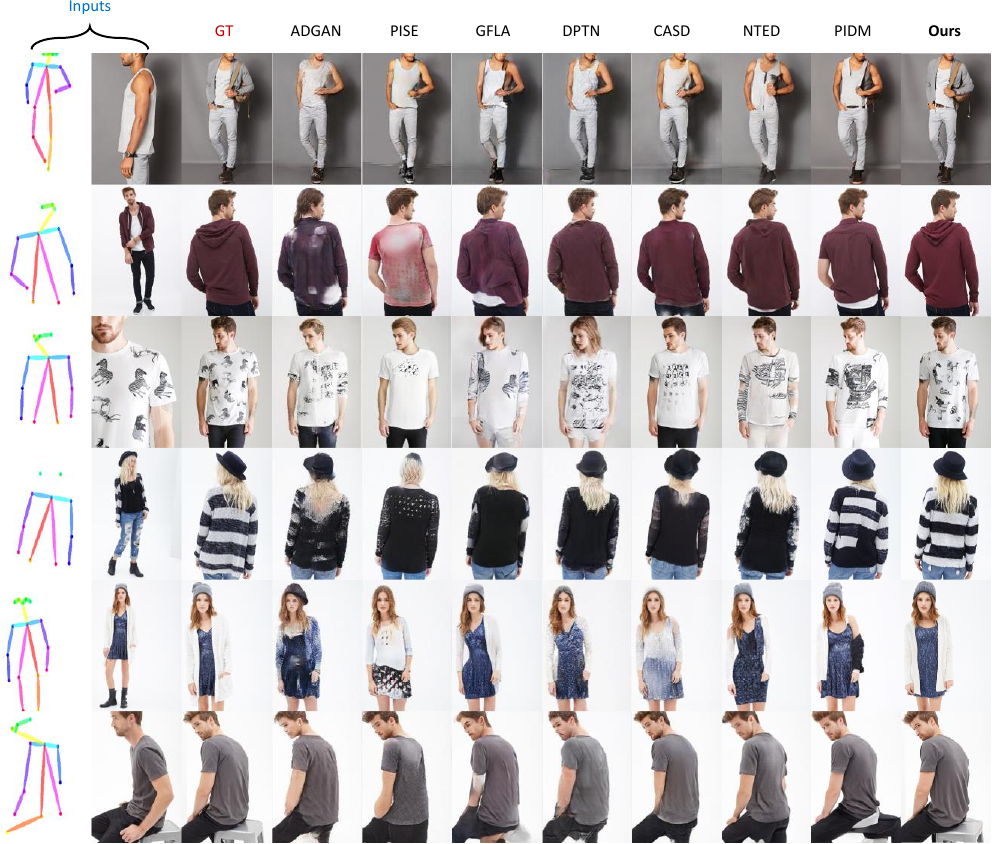}
  \vspace{-6pt}
  \caption{ More comparison images with state-of-the-art approaches. X-MDPT generates a target image that is more comprehensive and closely aligns with the ground truth image. Images at $256\times 256$ resolution are resized to $256\times 176$. (2) }
  \label{fig:single3}
\end{figure*}

\begin{figure*}[!htbp]
  \centering
  \includegraphics[width=1.0\linewidth]{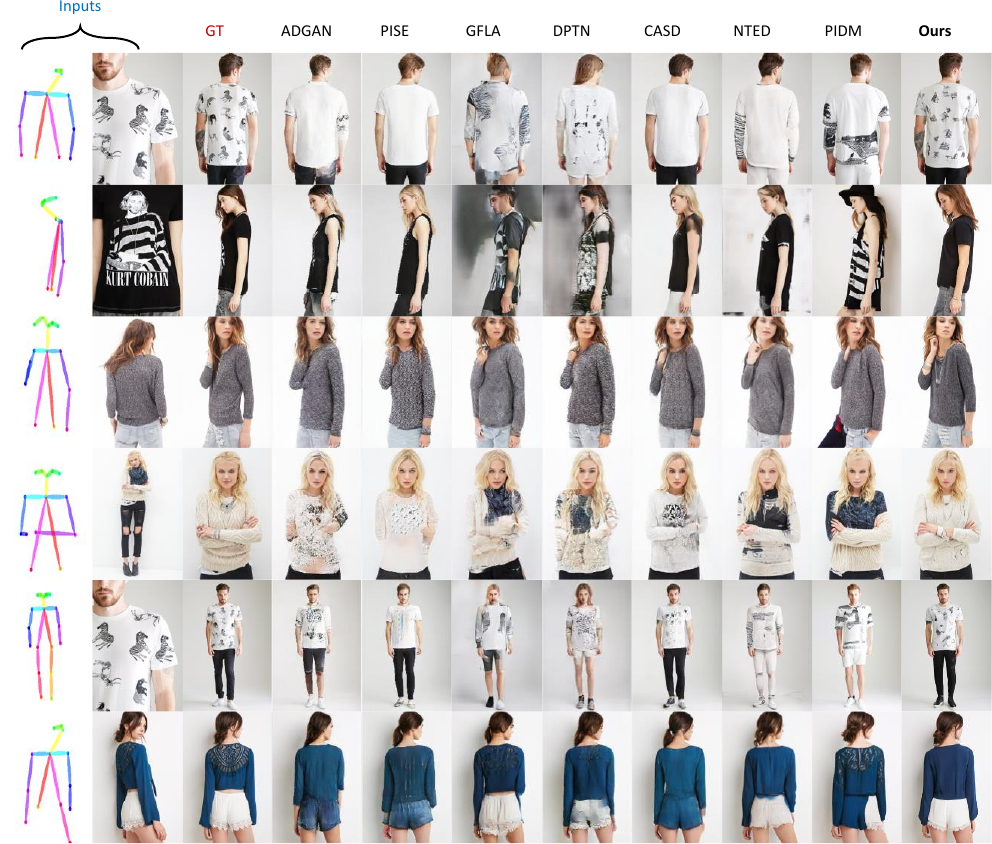}
  \vspace{-6pt}
  \caption{ More comparison images with state-of-the-art approaches. X-MDPT generates a target image that is more comprehensive and closely aligns with the ground truth image. Images at $256\times 256$ resolution are resized to $256\times 176$. (3) }
  \label{fig:single4}
\end{figure*}

\begin{figure*}[!htbp]
  \centering
  \includegraphics[width=1.0\linewidth]{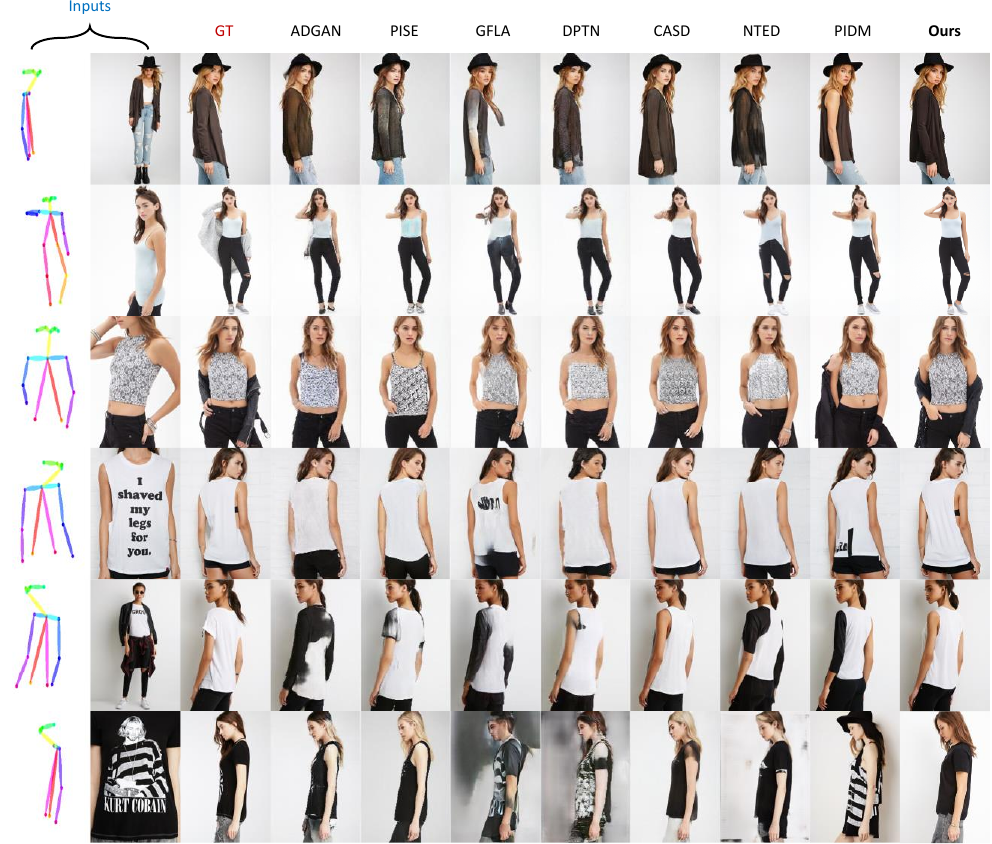}
  \vspace{-6pt}
  \caption{ More comparison images with state-of-the-art approaches. X-MDPT generates a target image that is more comprehensive and closely aligns with the ground truth image. Images at $256\times 256$ resolution are resized to $256\times 176$. (4) }
  \label{fig:single5}
\end{figure*}

\begin{figure*}[!htbp]
  \centering
  \includegraphics[width=1\linewidth]{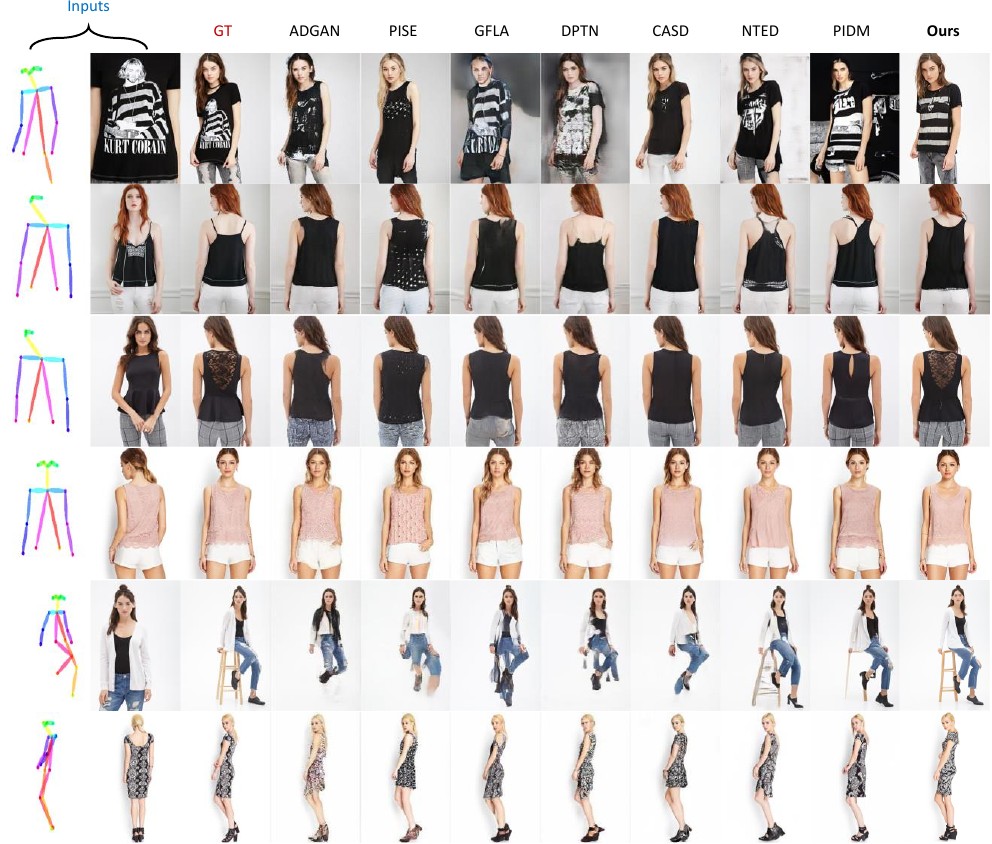}
  \vspace{-6pt}
  \caption{ More comparison images with state-of-the-art approaches. X-MDPT generates a target image that is more comprehensive and closely aligns with the ground truth image. Images at $256\times 256$ resolution are resized to $256\times 176$. (5) }
  \label{fig:single6}
\end{figure*}

\end{document}